%% file: main.tex
\documentclass[10pt,twocolumn,letterpaper]{article}

\usepackage{iccv}              % To produce the CAMERA-READY version

\definecolor{iccvblue}{rgb}{0.21,0.49,0.74}
\usepackage[pagebackref,breaklinks,colorlinks,allcolors=iccvblue]{hyperref}
\usepackage{lipsum}
\usepackage{svg}
\usepackage{hyperref}
\usepackage{url}
\usepackage{booktabs}
\usepackage{amssymb}% http://ctan.org/pkg/amssymb
\usepackage{pifont}% http://ctan.org/pkg/pifont
\usepackage{multirow}
\usepackage{multicol}
\usepackage{nicefrac}
\usepackage{bm}
\usepackage{graphicx}
\usepackage{colortbl}
\usepackage{amsmath, amssymb, amsthm}
\usepackage{booktabs}
\usepackage{siunitx}
\usepackage{graphicx}
\usepackage{xcolor}
\usepackage{multicol}
\usepackage{colortbl}
\usepackage{tabularx}
\usepackage{makecell}
\usepackage{multirow}
\usepackage{diagbox}
\usepackage{wrapfig}
\usepackage{enumitem}
\usepackage{algorithm, algorithmic}
\usepackage[normalem]{ulem}
\usepackage{tgcursor}
\usepackage{bbm}
\usepackage[utf8]{inputenc}
\usepackage{amsmath}
\usepackage{amssymb}
\usepackage{url}
\usepackage{enumitem}
\usepackage{caption}
\usepackage{tocloft}
\usepackage{pifont}

\usepackage{svg}
\usepackage{amsthm}
\usepackage{tablefootnote}
\usepackage{threeparttable}
\usepackage{float}
\newtheorem{assumption}{Assumption}

\def\paperID{4428} % *** Enter the Paper ID here
\def\confName{ICCV}
\def\confYear{2025}

\title{From Reusing to Forecasting: Accelerating Diffusion Models with TaylorSeers}

\vspace{-3mm}

\author{Jiacheng Liu$^{1,2}
\thanks{Equal contribution. shenyizou@outlook.com \\ 
$^\dag$Corresponding author: zhanglinfeng@sjtu.edu.cn}$,~
Chang Zou$^{1,3*}$, ~
Yuanhuiyi Lyu$^4$, ~
Junjie Chen$^1$,    ~
Linfeng Zhang$^{1\dag}$ \\\\
$^1$Shanghai Jiao Tong University ~~
$^2$Shandong University \\
$^3$University of Electronic Science and Technology of China\\
$^4$The Hong Kong University of Science and Technology (Guangzhou) \\
}

\begin{document}
\maketitle
\input{sec/0_abstract}
\input{sec/1_intro}
\begin{figure*}[t]
\centering
\includegraphics[width=\linewidth]{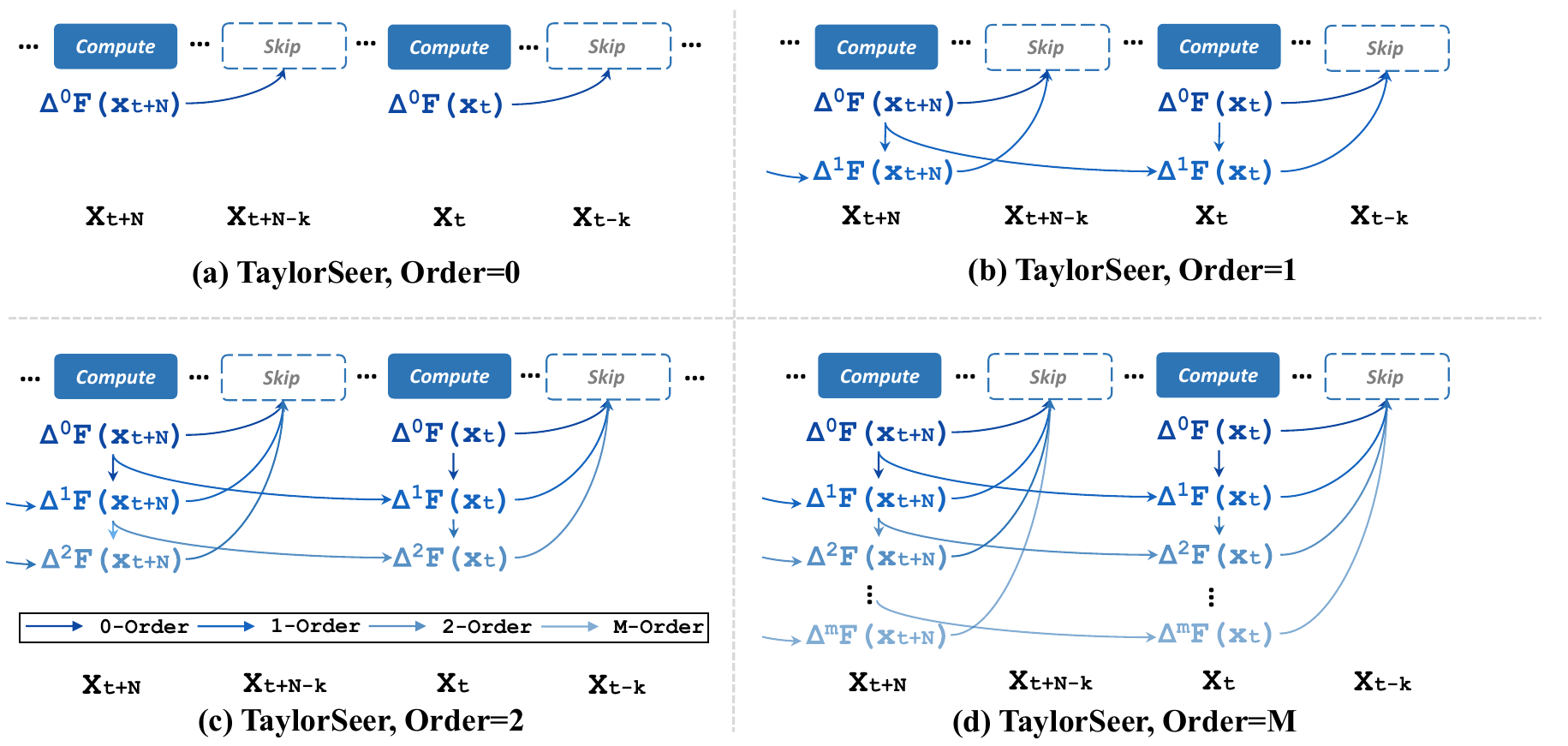}
\vspace{-8mm}
\caption{ \textbf{An overview of TaylorSeer}.$\quad$(a) \textbf{TaylorSeer (Order=0)} \textit{Naïve feature caching}, which directly reuses computed features across timesteps.$\quad$(b)  \textbf{TaylorSeer (Order=1)}  \textit{Linear prediction}, which estimates feature trajectories using first-order finite differences.$\quad$(c)  \textbf{TaylorSeer (Order=2)} extends linear prediction to 2-order finite differences for more accurate modeling of nonlinear feature trajectories. (d) \textbf{TaylorSeer (Order=M)} further extends to M-orders for improved accuracy without sacrificing efficiency.
}\vspace{-5mm}
\label{fig:Method}
\end{figure*}
\input{sec/2_related_works}
\input{sec/3_method}
\input{sec/4_experiments}
\input{sec/5_conclusion}

{
    \small
    \bibliographystyle{ieeenat_fullname}
    \bibliography{main}
}
\input{sec/X_suppl}

\end{document}

%% file: sec/0_abstract.tex
\begin{abstract}
Diffusion Transformers (DiT) have revolutionized high-fidelity image and video synthesis, yet their computational demands remain prohibitive for real-time applications.
To solve this problem, feature caching has been proposed to accelerate diffusion models by caching the features in the previous timesteps and then reusing them in the following timesteps.
However, at timesteps with significant intervals, the feature similarity in diffusion models decreases substantially, leading to a pronounced increase in errors introduced by feature caching, significantly harming the generation quality.
To solve this problem, we propose TaylorSeer, which firstly shows that features of diffusion models at future timesteps can be predicted based on their values at previous timesteps.
Based on the fact that features change slowly and continuously across timesteps, 
TaylorSeer employs a differential method to approximate the higher-order derivatives of features and predict features in future timesteps with Taylor series expansion. 
Extensive experiments demonstrate its significant effectiveness in both image and video synthesis, especially in high acceleration ratios.
For instance, it achieves an almost lossless acceleration of 4.99$\times$ on FLUX and 5.00$\times$ on HunyuanVideo without additional training. On DiT, it achieves $3.41$ lower FID compared with previous SOTA at $4.53$$\times$ acceleration.
%Our code is provided in the supplementary materials and will be made publicly available on GitHub.
\\Our codes have been released in Github: \\ 
\textbf{\href{https://github.com/Shenyi-Z/TaylorSeer}{\texttt{\textcolor{cyan}{https://github.com/Shenyi-Z/TaylorSeer}}}}
\end{abstract}

%% file: sec/1_intro.tex
\vspace{-3mm}
\section{Introduction} \label{sec:intro}
%\vspace{-2mm}
Diffusion Models (DMs)~\cite{DM} have made significant strides in generative artificial intelligence, achieving remarkable results in tasks such as image generation and video synthesis~\cite{StableDiffusion,blattmann2023SVD}.
The introduction of Diffusion Transformers (DiT)~\cite{DiT} has further advanced the quality of visual generation. However, these improvements come with a substantial increase in computational demands, limiting the practical use of diffusion transformers. In response to these growing challenges regarding computational efficiency, 
various acceleration techniques have been proposed ~\cite{yuan2024ditfastattn,zou2024accelerating, ma2024deepcache,zhao2024PAB}. 
\begin{figure}
\centering\includegraphics[width=\linewidth]{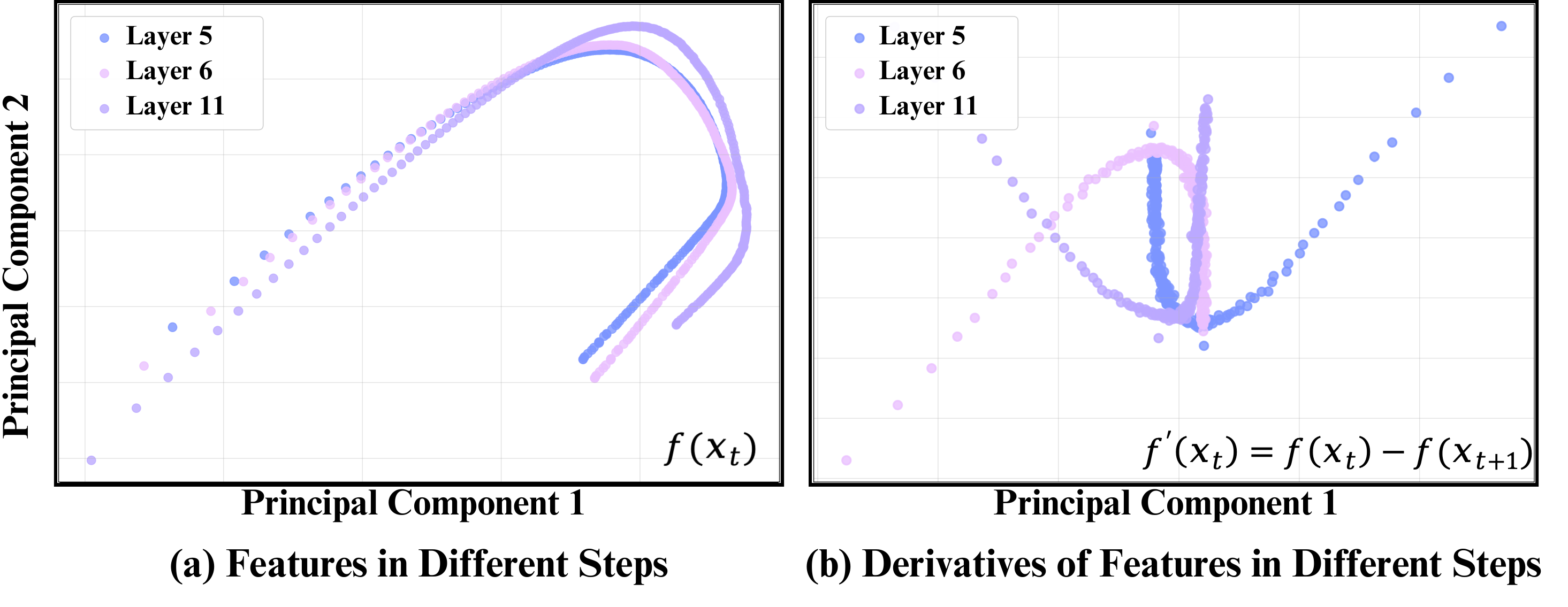}
\vspace{-8mm}
\caption{\textbf{PCA projections of features and their derivatives in diffusion models.} Both the features (a) and derivatives of features (b) in diffusion models at different timesteps form stable trajectories, indicating it possible to predict features of diffusion models at future timesteps based on features from previous timesteps.
}
\vspace{-7mm}

\label{fig:DiT_Feature}
\end{figure}
Recently, based on the observation that diffusion models have highly similar features in the adjacent timesteps, feature caching methods have been proposed to store the features in the previous timesteps and then reuse them in the following timesteps, allowing diffusion models to skip substantial computations in both U-Net-based and transformer-based 
diffusion models~\cite{ma2024deepcache,selvaraju2024fora,chen2024delta-dit} without requirements for additional training.
Previous caching methods follow the \textit{``cache-then-reuse''} paradigm and explore it from the perspectives of tokens~\cite{zou2024accelerating,zou2024DuCa,selvaraju2024fora} and residuals~\cite{chen2024delta-dit} while ignoring the its natural limitation: \emph{
As the distance between two timesteps increases, their feature similarity decreases exponentially, making reusing features at distant steps significantly harm the generation quality}.
This curse locks the possibility of feature caching in high-ratio acceleration. As shown in the PCA results from Figure~\ref{fig:DiT_Feature}(a), features of diffusion models in non-adjacent timesteps exhibit long distances, verifying this limitation and calling for the development of a new caching paradigm.

\begin{figure}
\centering
\includegraphics[width=0.95\linewidth]{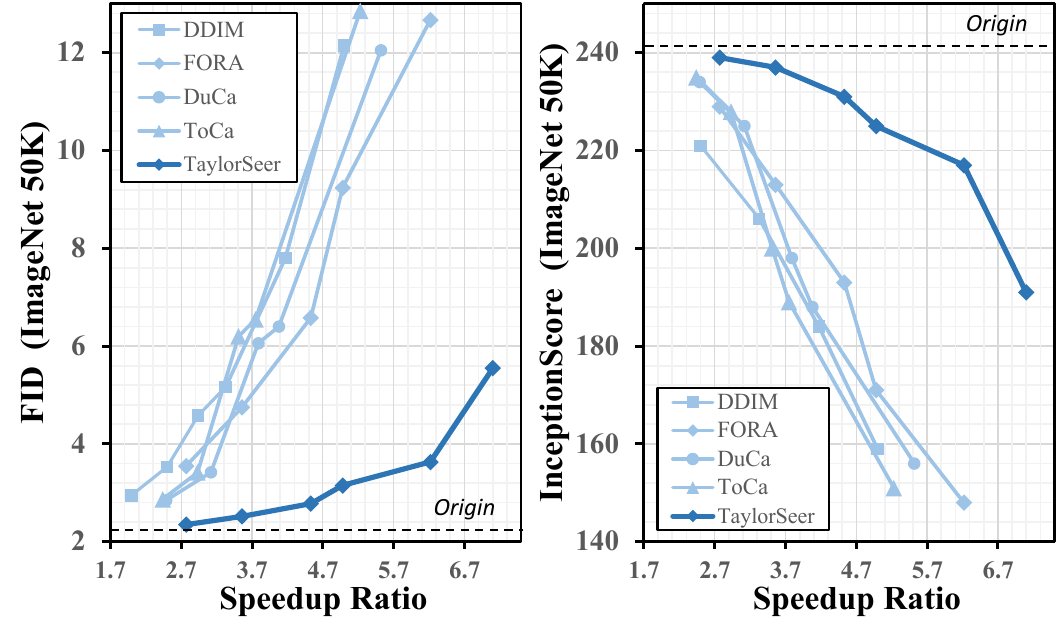}
\caption{\textbf{Comparison between previous caching methods and TaylorSeer.} TaylorSeer shows significantly better performance at high-acceleration ratios.}
\label{fig:fid-flops}
\vspace{-7mm}

\end{figure}

To solve this problem, this paper introduces a new paradigm, \textit{``cache-then-forecast''}, to replace the previous \textit{``cache-then-reuse''}.
As shown in Figure~\ref{fig:DiT_Feature}(a),  features of the diffusion model at different timesteps form a stable trajectory, demonstrating that it is possible to predict the features at the future timesteps based on features from previous timesteps.
Furthermore, we investigate the derivatives of features in different timesteps (\emph{i.e.,} the velocity along the trajectory) and present their PCA results in Figure~\ref{fig:DiT_Feature}(b).
Surprisingly, derivatives of features also exhibit similar values at the adjacent timesteps with high stability, indicating that predicting future features is not a complex problem and may be solvable using a non-parametric method.

Building on this observation, we propose \textit{TaylorSeer}, a \textit{``cache-then-forecast''} paradigm that leverages Taylor series expansion to predict the features at future timesteps.
Specifically, \textit{TaylorSeer} utilizes multi-step features to approximate derivatives of various orders as features evolve over timesteps using difference methods and applies the Taylor series to predict subsequent features.
Unlike previous methods that directly reuse cached features, our approach exploits the continuity of feature changes to predict future features, allowing diffusion models to achieve a training-free and high-ratio acceleration without significant decrements in generation quality.
Compared with previous caching methods that suffer from low feature similarity at distant timesteps, \textit{TaylorSeer} is particularly effective at larger intervals between full activations, where higher-order Taylor series approximations excel in long-range feature reuse scenarios.
As shown in Figure~\ref{fig:fid-flops}, our method reduces quality loss compared with the previous SOTA by 36$\times$, achieving promising performance in acceleration regimes beyond 6$\times$, where all previous methods fail.

In summary, our contributions are as follows:
\begin{itemize}
\item \textbf{Cache-then-forecast Paradigm}: We propose a novel paradigm of \textit{``cache-then-forecast''} to replace the previous \textit{``cache-then-resue''}, enabling the prediction of diffusion model features at future timesteps through sequential modeling and overcoming the limitations of caching-based methods in high-acceleration regimes.
\item \textbf{TaylorSeer}: We introduce \textit{TaylorSeer}, which utilizes Taylor series expansion to approximate the trajectory of features at different timesteps using higher-order derivatives, surpassing previous methods by a large margin without introducing any training or search costs.
\item \textbf{State-of-the-Art Performance}: \textit{TaylorSeer} achieves \textbf{2.5$\times$}, \textbf{4.99$\times$}, and \textbf{5.00$\times$} acceleration on DiT, FLUX, and HunyuanVideo for image and video synthesis, respectively, while maintaining high-quality generation 
and sometimes even providing additional benefits, paving a new path for diffusion model acceleration.
\end{itemize}

%% file: sec/2_related_works.tex
\section{Related Works}\label{sec:Related Works}
\vspace{-1mm}

Diffusion models~\cite{sohl2015deep,ho2020DDPM} have demonstrated exceptional capabilities in image and video generation. Initial architectures primarily utilized U-Net-based designs~\cite{ronneberger2015unet} which, despite their efficacy, encountered scalability constraints that limited larger model training and practical deployment. The introduction of Diffusion Transformer (DiT)~\cite{peebles2023dit} addressed these limitations, subsequently inspiring numerous advancements that adapted the architecture to achieve state-of-the-art performance across diverse domains~\cite{chen2023pixartalpha,chen2024pixartsigma,opensora,yang2025cogvideox}. Despite these achievements, the inherent sequential sampling process of diffusion models imposes substantial computational demands during inference. Consequently, acceleration techniques have emerged as a critical research focus, broadly categorized as \textit{Sampling Timestep Reduction} and \textit{Denoising Network Acceleration}.

\vspace{-1mm}
\subsection{Sampling Timestep Reduction}
\vspace{-1mm}

A fundamental approach to accelerating diffusion models involves \textit{minimizing sampling steps while preserving output quality}. DDIM~\cite{songDDIM} established a deterministic sampling methodology that maintained generation fidelity with reduced denoising iterations. The DPM-Solver series~\cite{lu2022dpm,lu2022dpm++,zheng2023dpmsolvervF} further advanced this concept through high-order ODE solvers. Alternative strategies include Rectified Flow~\cite{refitiedflow}, which constructs direct paths between noise and data distributions, and knowledge distillation techniques~\cite{salimans2022progressive,meng2022on} that compress multiple denoising operations into fewer steps. Notably, Consistency Models~\cite{song2023consistency} introduced an innovative framework enabling single-step or few-step sampling by directly mapping noisy inputs to clean data, eliminating sequential denoising requirements and significantly enhancing practical applicability.

\vspace{-1mm}
\subsection{Denoising Network Acceleration}
\vspace{-1mm}

In addition to reducing the number of sampling timesteps, \textit{optimizing the computational efficiency of the denoising network itself} presents another promising approach for accelerating inference. This can be broadly classified into \textit{Model Compression-based} and \textit{Feature Caching-based} techniques, as detailed below.
\vspace{-3mm}

\paragraph{Model Compression-based Acceleration.}
Model compression techniques encompass network pruning~\cite{structural_pruning_diffusion, zhu2024dipgo}, quantization~\cite{10377259, shang2023post, kim2025ditto}, knowledge distillation~\cite{li2024snapfusion}, and token reduction~\cite{bolya2023tomesd, kim2024tofu, zhang2024tokenpruningcachingbetter, zhang2025sito, cheng2025catpruningclusterawaretoken, saghatchian2025cached}.
These methods typically require additional training to fine-tune the compressed model, ensuring minimal degradation in generation quality while significantly improving inference speed.
Despite their effectiveness, these approaches often involve a trade-off between model size reduction and potential loss of expressive power, making it essential to design efficient compression strategies tailored to diffusion models.
\vspace{-5mm}
\paragraph{Feature Caching-based Acceleration.}
Feature caching has emerged as particularly advantageous due to its training-free implementation. Initially developed for U-Net architectures~\cite{li2023FasterDiffusion, ma2024deepcache}, caching mechanisms have evolved to address the computational demands of DiT models, which offer superior performance but at increased computational cost. Advanced techniques such as FORA~\cite{selvaraju2024fora} and $\Delta$-DiT~\cite{chen2024delta-dit} leverage attention and MLP representation reuse, while TeaCache~\cite{liu2024timestep} implements dynamic timestep-dependent difference estimation to optimize caching decisions. DiTFastAttn~\cite{yuan2024ditfastattn} reduces redundancies in self-attention computation across spatial, temporal, and conditional dimensions. The ToCa series~\cite{zou2024accelerating, zou2024DuCa, Liu2025SpeCa, Zheng2025Compute} mitigates information loss through dynamic feature updates, while EOC~\cite{qiu2025acceleratingdiffusiontransformererroroptimized} introduces an error-optimized framework utilizing prior knowledge extraction and adaptive optimization. FasterCache~\cite{lvFasterCacheTrainingFreeVideo2025} extends caching to video DiTs through dynamic feature reuse and a CFG-Cache strategy. Recent innovations including UniCP~\cite{sun2025unicpunifiedcachingpruning}, which integrates dynamic cache window adjustment with pruning, and RAS~\cite{liu2025regionadaptivesamplingdiffusiontransformers}, which implements region-adaptive sampling rates based on model focus, further enhance computational efficiency while preserving generation fidelity.  Caching has also been widely adopted in other tasks such as image editing~\cite{yanEEditRethinkingSpatial2025}.

Despite these advancements, existing caching methods predominantly follow a \textit{``cache-then-reuse''} paradigm, which stores and reuses features based on tokens and residuals. However, as the timestep gap increases, feature similarity decreases exponentially, leading to degradation in generation quality and limiting the potential acceleration gains.
In this work, we focus on advancing feature caching techniques by introducing a novel \textit{``cache-then-forecast''} paradigm.
Instead of directly reusing stored features, our approach predicts future features based on past ones, leveraging their stable trajectory across timesteps.
This predictive strategy enables significantly higher acceleration while maintaining generation fidelity, all without relying on complex additional models.
\vspace{-2mm}

%% file: sec/3_method.tex
\section{Method}
\vspace{-2mm}
\subsection{Preliminary}
\vspace{-1mm}
\paragraph{Diffusion Models.}  
Diffusion models operate through a continuous-time stochastic process with forward and reverse phases for precise noise addition and removal, governed by stochastic differential equations (SDEs). The forward diffusion process is mathematically defined as:  
\begin{equation}
dx = \sqrt{\frac{d\sigma^2(t)}{dt}}dw,
\end{equation}
where $\sigma(t)$ denotes the noise schedule, and $w$ represents the standard Wiener process. The corresponding reverse process restores data by removing noise:  
\begin{equation}
dx = \left[-\frac{1}{2} \beta(t) x - \nabla_{x_t} \log p_t(x_t)\right] dt + \sqrt{\beta(t)} d\bar{w}.
\end{equation}  
This formulation ensures that features evolve smoothly over time in a continuous manner. In practical implementations, the process is discretized into timesteps, maintaining its structured trajectory. Specifically, the denoising transition at each step follows a Gaussian distribution:
\begin{equation}
p_\theta(x_{t-1} | x_t) = \mathcal{N} \left(x_{t-1}; \mu_\theta(x_t, t), \beta_t \mathbf{I} \right).
\end{equation}
Despite the inherent discretization challenges, the underlying feature transformations remain predictable due to the fundamentally smooth nature of the diffusion trajectory.

\begin{assumption}  
The feature representations in diffusion models evolve smoothly over time. Specifically, the underlying feature transformation is a differentiable function with bounded higher-order derivatives, ensuring structured and predictable variation across timesteps. This smoothness persists under numerical discretization, providing a foundation for feature caching strategies.  \vspace{-3mm}
\end{assumption}
\vspace{-2mm}
\paragraph{Diffusion Transformer Architecture.}
The Diffusion Transformer (DiT) follows a hierarchical structure, $\mathcal{G} = g_1 \circ g_2 \circ \cdots \circ g_L$, where each block $g_l = \mathcal{F}_{\text{SA}}^l \circ \mathcal{F}_{\text{CA}}^l \circ \mathcal{F}_{\text{MLP}}^l$ consists of self-attention (SA), cross-attention (CA), and multilayer perceptron (MLP) components. The superscript $l \in \{1,2,...,L\}$ denotes the layer index. In DiT, both attention mechanisms and MLP components evolve over time. At each timestep, $\mathcal{F}_{\text{SA}}^l$, $\mathcal{F}_{\text{CA}}^l$, and $\mathcal{F}_{\text{MLP}}^l$ dynamically adjust to accommodate varying noise levels during image generation. The input, $\mathbf{x}_t = \{x_i\}_{i=1}^{H \times W}$, is represented as a sequence of tokens corresponding to image patches. Each block integrates residual connections, expressed as $\mathcal{F}(\mathbf{x}) = \mathbf{x} + \text{AdaLN} \circ f(\mathbf{x})$, where $f(\mathbf{x})$ corresponds to MLP or attention layers, and AdaLN denotes adaptive layer normalization, ensuring consistent information flow.
\vspace{-3mm}
\paragraph{Na\"ive Feature Caching for Diffusion Models.}
Recent acceleration methods employ \textit{Na\"ive Feature Caching Strategies} in diffusion models by directly reusing computed features across adjacent timesteps.
Specifically, given timesteps $\{t,t-1,\dots,t-(N-1)\}$, features computed at timestep $t$ are cached as $\mathcal{C}(x_t^l) := \{\mathcal{F}(x_t^l) \}$.
These cached features are then directly reused for subsequent steps: $\mathcal{F}(x_{t-k}^l) := \mathcal{F}(x_t^l)$, where $k \in {1, \dots, N-1}$.
While this approach achieves a theoretical $(N-1)$-fold speedup by eliminating redundant computations, it suffers from exponential error accumulation as $N$ increases due to neglecting the temporal dynamics of features.

\subsection{TaylorSeer}  
\textit{TaylorSeer} introduces Taylor series-based predictive caching to mitigate error accumulation in conventional caching approaches. By leveraging the continuous nature of feature trajectory, we develop a predictive caching method that accurately estimates intermediate features across timesteps.
\vspace{-3mm}
\paragraph{Linear Prediction Method.}
To overcome the limitations of \textit{naïve caching}, we propose a \textit{linear prediction strategy} that extends beyond direct feature reuse.
Instead of directly copying features, we cache both feature values and their temporal differences: $\mathcal{C}(x_t^l) := \{\mathcal{F}(x_t^l), \Delta\mathcal{F}(x_t^l)\}$.
This allows us to predict feature trajectories at timestep $t-k$ using the following formulation:
\begin{equation}
\mathcal{F}_{\textrm{pred}}(x_{t-k}^l) = \mathcal{F}(x_t^l) + \frac{\mathcal{F}(x_t^l) - \mathcal{F}(x_{t+N}^l)}{N}k
\end{equation}
where $\mathcal{F}(x_{t+N}^l)$ and $\mathcal{F}(x_t^l)$ denote features at fully activated timesteps. This first-order approximation captures the linear trend of feature trajectories, significantly improving accuracy over direct feature reuse.
\vspace{-3mm}
\paragraph{Higher-Order Prediction via Taylor Expansion.}  
To further improve the accuracy of feature prediction, we extend our forecasting method by incorporating Taylor's theorem.
This approach leverages higher-order finite difference approximations to capture the temporal dynamics of features, reducing cumulative prediction errors while maintaining computational efficiency.
At timestep $t$, we define a cache storing the feature and its $m$-th order finite differences:
\begin{equation}
\mathcal{C}(x_t^l) := \{\mathcal{F}(x_t^l), \Delta\mathcal{F}(x_t^l), ..., \Delta^m\mathcal{F}(x_t^l)\}
\end{equation}
where $\Delta^i \mathcal{F}(x_t^l)$ represents the $i$-th order finite difference, approximating the temporal dynamics of $\mathcal{F}(x_t^l)$ around timestep $t$. For an $(m+1)$-times differentiable feature function $\mathcal{F}(x_t^l)$, the feature at timestep $t-k$ can be expressed using a Taylor series expansion:
\begin{equation}
\mathcal{F}(x_{t-k}^l) = \mathcal{F}(x_t^l) + \sum_{i=1}^{m} \frac{\mathcal{F}^{(i)}(x_t^l)}{i!}(-k)^i + R_{m+1}
\end{equation}
where $R_{m+1}$ represents the remainder term.
To avoid explicit computation of higher-order derivatives, we approximate them using finite differences. The $i$-th order forward finite difference is defined recursively as:
\begin{equation}
\Delta^i \mathcal{F}(x_t^l) = \Delta(\Delta^{i-1}\mathcal{F}(x_t^l)) = \Delta^{i-1}\mathcal{F}(x_{t+N}^l) - \Delta^{i-1}\mathcal{F}(x_t^l)
\label{eq: difference-approximation}
\end{equation}
with the base case $\Delta^0\mathcal{F}(x_t^l) = \mathcal{F}(x_t^l)$. Equation~(\ref{eq: difference-approximation}) can be further expanded into a binomial form:
\begin{equation}
\Delta^i \mathcal{F}(x_t^l) = \sum_{j=0}^{i} (-1)^{i-j} \binom{i}{j} \mathcal{F}(x_{t+jN}^l)
\end{equation}
It can be shown that the $i$-th order finite difference approximates the $i$-th derivative scaled by $N^i$:
\begin{equation}
\Delta^i\mathcal{F}(x_t^l) \approx N^i \mathcal{F}^{(i)}(x_t^l)
\end{equation}
Substituting this approximation into the Taylor expansion and accounting for the scaling factor, we derive the $m$-th order prediction formula:
\begin{equation}
\mathcal{F}_{\textrm{pred},m}(x_{t-k}^l) = \mathcal{F}(x_t^l) + \sum_{i=1}^{m} \frac{\Delta^i\mathcal{F}(x_t^l)}{i! \cdot N^i}(-k)^i
\end{equation}
This formulation requires only $(m+1)$ fully computed timesteps $\{t+mN, \dots, t+N, t\}$ to predict features at intermediate timesteps, achieving an optimal balance between efficiency and accuracy.

Our method establishes a principled transition from directly feature caching to predictive forecasting by leveraging the mathematical foundation of Taylor series approximation. Instead of merely reusing features, we model their temporal trajectory over time, enabling accurate estimation of intermediate representations and capturing both short-term and long-term dynamics. This systematic formulation unifies various prediction strategies—ranging from simple caching to higher-order forecasting—within a single coherent methodology, providing flexibility and robustness across different temporal scales and scenarios.
\begin{itemize}
\item  \textbf{Directly Caching ($m=0$)}: Reduces to \textit{Naïve Feature Caching}, reusing features without temporal modeling. Simple but less accurate for dynamic feature changes.
\item \textbf{Short-Term Forecasting ($m=1$)}: Uses first-order finite differences for linear prediction, suitable for short-term changes but limited for complex dynamics.
\item \textbf{Long-Range Forecasting ($m \geq 2$)}: Leverages higher-order finite differences to model nonlinear trajectories, reducing errors and capturing long-term temporal patterns.
\end{itemize}
\vspace{1mm}
Unlike Naïve Feature Caching,  \textit{TaylorSeer} transforms simply feature reuse into a predictive process, explicitly modeling feature trajectories through high-order finite differences. This key transition from directly caching to forecasting not only significantly 
 improves prediction accuracy but also enables efficient long-range inference. As a result, \textit{TaylorSeer} provides a robust and scalable solution for accelerating diffusion model inference, maintaining high generation quality even at large acceleration ratios.
\vspace{-3mm}
\paragraph{Error Bounds Analysis.}
For a feature function $\mathcal{F}(x_t^l)$ that is $(m+1)$-times differentiable on $[t-k,t]$, we define the prediction error as $E_m(k) = \|\mathcal{F}_{\textrm{pred},m}(x_{t-k}^l) - \mathcal{F}(x_{t-k}^l)\|$. By Taylor's remainder theorem, this error is bounded by:
\begin{equation}
E_m(k) \leq \frac{M_{m+1}}{(m+1)!}|k|^{m+1}
\end{equation}
where $M_{m+1} = \sup_{\xi \in [t-k,t]} \|\mathcal{F}^{(m+1)}(x_\xi^l)\|$. Since our method employs finite difference approximations rather than exact derivatives, the complete error bound incorporates additional terms:
\begin{equation}
E_m(k) \leq \frac{M_{m+1}}{(m+1)!}|k|^{m+1} + \sum_{i=1}^{m} \frac{C_i}{i!\cdot|N|^{i-1}}|k|^i
\end{equation}
where constants $C_i$ relate to finite difference approximation errors. This analysis reveals a fundamental trade-off: higher-order predictions ($m$) effectively reduce the primary error term but introduce additional errors scaling with the sampling interval $N$. For diffusion models with smooth feature trajectories (small $M_{m+1}$), our method achieves optimal accuracy by balancing prediction order and caching interval, particularly effective when $|k| < |N|$ where the Taylor approximation dominates the error bound.

%% file: sec/4_experiments.tex
\section{Experiments}
\input{tab/FLUX-Metrics}
\input{tab/HunyuanVideo-Metrics}

\subsection{Experiment Settings}
\paragraph{Model Configurations.}
The experiments are conducted on three state-of-the-art visual generative models: the text-to-image generation model FLUX.1-dev~\citep{flux2024}, text-to-video generation model HunyuanVideo~\citep{sun_hunyuan-large_2024}, and the class-conditional image generation model DiT-XL/2~\citep{DiT}. 

\noindent \textbf{FLUX.1-dev}\citep{flux2024} predominantly employs the Rectified Flow~\citep{refitiedflow} sampling method with 50 steps as the standard configuration. All experimental evaluations of FLUX.1-dev were conducted on NVIDIA H800 GPUs.

\noindent \textbf{HunyuanVideo}~\citep{li_hunyuan-dit_2024, sun_hunyuan-large_2024} was evaluated on the Hunyuan-Large architecture, utilizing the standard 50-step inference protocol as the baseline while preserving all default sampling parameters for rigorous experimental consistency. Extensive performance benchmarks were systematically conducted using NVIDIA H20 96GB GPUs for detailed latency assessment and NVIDIA H100 80GB GPUs for comprehensive inference operations.

\noindent \textbf{DiT-XL/2}~\cite{DiT} adopts a 50-step DDIM~\citep{songDDIM} sampling strategy to ensure consistency with other models. All models incorporate a unified forced activation period $\mathcal{N}$, while $\mathcal{O}$ represents the order of the Taylor expansion, optimizing computational efficiency and overall model performance. Experiments on DiT-XL/2 are conducted on NVIDIA A800 80GB GPUs.
\textit{For more detailed model configurations, please refer to the Supplementary Material.}

\vspace{-5mm}
\paragraph{Evaluation and Metrics.}
For the text-to-image generation task, we perform inference on 200 DrawBench~\cite{sahariaPhotorealisticTexttoImageDiffusion2022} prompts to generate images with a resolution of 1000 $\times$ 1000. We then evaluate the generated samples using ImageReward~\cite{xuImageRewardLearningEvaluating2023} and CLIP Score~\cite{CLIPScore} as key metrics to assess image quality and text alignment. For the text-to-video generation task, we leverage the VBench~\cite{VBench} evaluation framework, utilizing its 946 benchmark prompts. For each prompt, we generate five samples with different random seeds, totaling 4,730 videos. We then systematically assess the generated results based on 16 core evaluation dimensions defined by the VBench framework to provide a comprehensive evaluation of the model's performance. For the class-conditional image generation task, we uniformly sample from 1,000 ImageNet~\cite{Imagenet} categories, generating 50,000 images with a resolution of 256 $\times$ 256. We use FID-50k~\cite{FID50K} as the primary evaluation metric, complemented by sFID and Inception Score for robust evaluation. 
We assess the fidelity of our accelerated outputs against the original results using PSNR, SSIM~\cite{wangImageQualityAssessment2004}, and LPIPS~\cite{zhangUnreasonableEffectivenessDeep2018}.

\vspace{-2mm}
\subsection{Text-to-Image Generation}
%\vspace{-1mm}

\paragraph{Quantitative Study.}
We compared \textit{TaylorSeer} with existing methods. While DuCa ($\mathcal{N}$=5) achieves a 3.45$\times$ speedup with an ImageReward of 0.9896 and ToCa ($\mathcal{N}$=5) offers 3.30$\times$ acceleration at reduced quality (0.9731), \textit{TaylorSeer} already demonstrates superior performance in this range. The performance gap widens significantly at higher accelerations. When pushed towards 5$\times$ speedup, baseline methods suffer severe quality degradation; for instance, DuCa ($\mathcal{N}$=6) and ToCa ($\mathcal{N}$=8) see their ImageReward drop to 0.9470 and 0.9086, respectively. In stark contrast, \textit{TaylorSeer} ($\mathcal{N}$=6, $\mathcal{O}$=2) sustains an exceptional 1.0039 ImageReward at a 4.99$\times$ acceleration, consistently outperforming all competitors in PSNR, SSIM, and LPIPS as well.

\vspace{-2mm}

\begin{figure}[h!]
\centering
\includegraphics[width=0.95\linewidth]{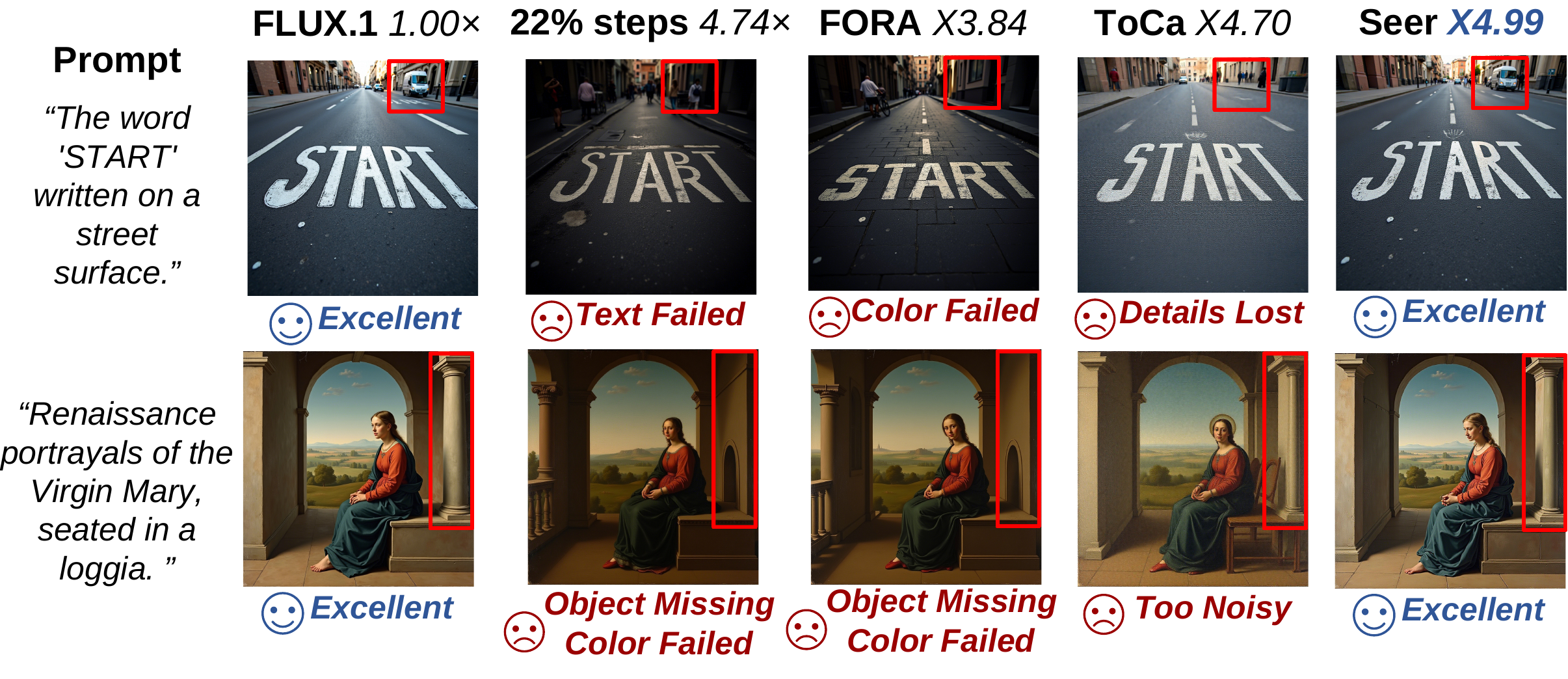}
\vspace{-3mm}
\caption{Detailed visualization results for different acceleration methods on FLUX.1-dev. Other methods exhibit issues such as text failure, color distortion, and missing details, whereas TaylorSeer achieves the best quality and acceleration.}
\label{fig:Vis-FLUX-small}
\vspace{-5mm}
\end{figure}

\input{tab/DiT-Metrics}
\paragraph{Qualitative Study.} The qualitative results highlight \textit{TaylorSeer}'s superior ability to preserve image quality while accelerating computation. In text generation tasks, such as \textit{``The word `START' written on a street surface,"} \textit{TaylorSeer} accurately retains the textual elements, whereas methods like FORA and ToCa lose critical details. In Renaissance-style portrait generation, \textit{TaylorSeer} maintains consistent image fidelity, while other methods exhibit notable \textit{color inaccuracies} and \textit{missing objects} across test cases. This shows \textit{TaylorSeer} balances efficiency and quality, particularly in tasks needing fine detail preservation.
% \vspace{-3mm}

\begin{figure*}[h!]
\centering
\includegraphics[width=\linewidth]{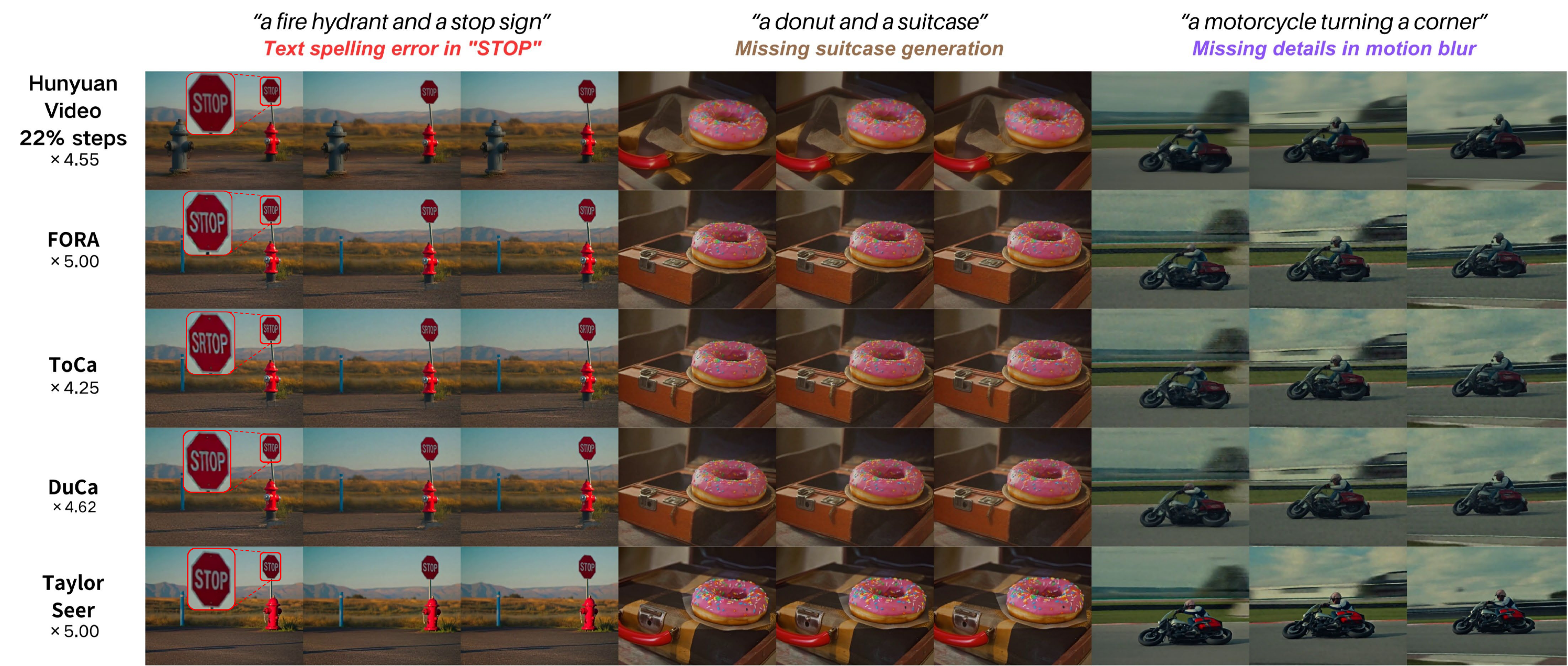}
\vspace{-6mm}
\caption{Visualization of different acceleration methods on HunyuanVideo. While achieving higher acceleration ratios, other methods exhibit issues such as \textit{text errors}, \textit{missing content}, and \textit{motion detail loss}. In contrast, our method demonstrates superior performance, maintaining high-quality generation without these problems.}
\label{fig:Vis-Huanyuan}
\vspace{-3mm}
\end{figure*}
 \vspace{-2mm}

\subsection{Text-to-Video Generation}
\vspace{-1mm}

\paragraph{Quantitative Study.}
On HunyuanVideo, \textit{TaylorSeer} with $\mathcal{N}$=5 and $\mathcal{O}$=1 reduces inference latency to 68 seconds and computational cost to 5960.4TFLOPs ($5.00\times$ speedup), achieving a 79.93\% VBench score that outperforms both ToCa and DuCa. With $\mathcal{N}$=6 and $\mathcal{O}$=1, performance improves further to $5.56\times$ speedup while maintaining a 79.78\% score. It also achieves closer-to-original outputs with better PSNR, SSIM, and LPIPS than all baselines.

\begin{figure*}[h!]
\centering
\begin{minipage}{0.28\linewidth}
\centering
\includegraphics[width=0.95\linewidth]{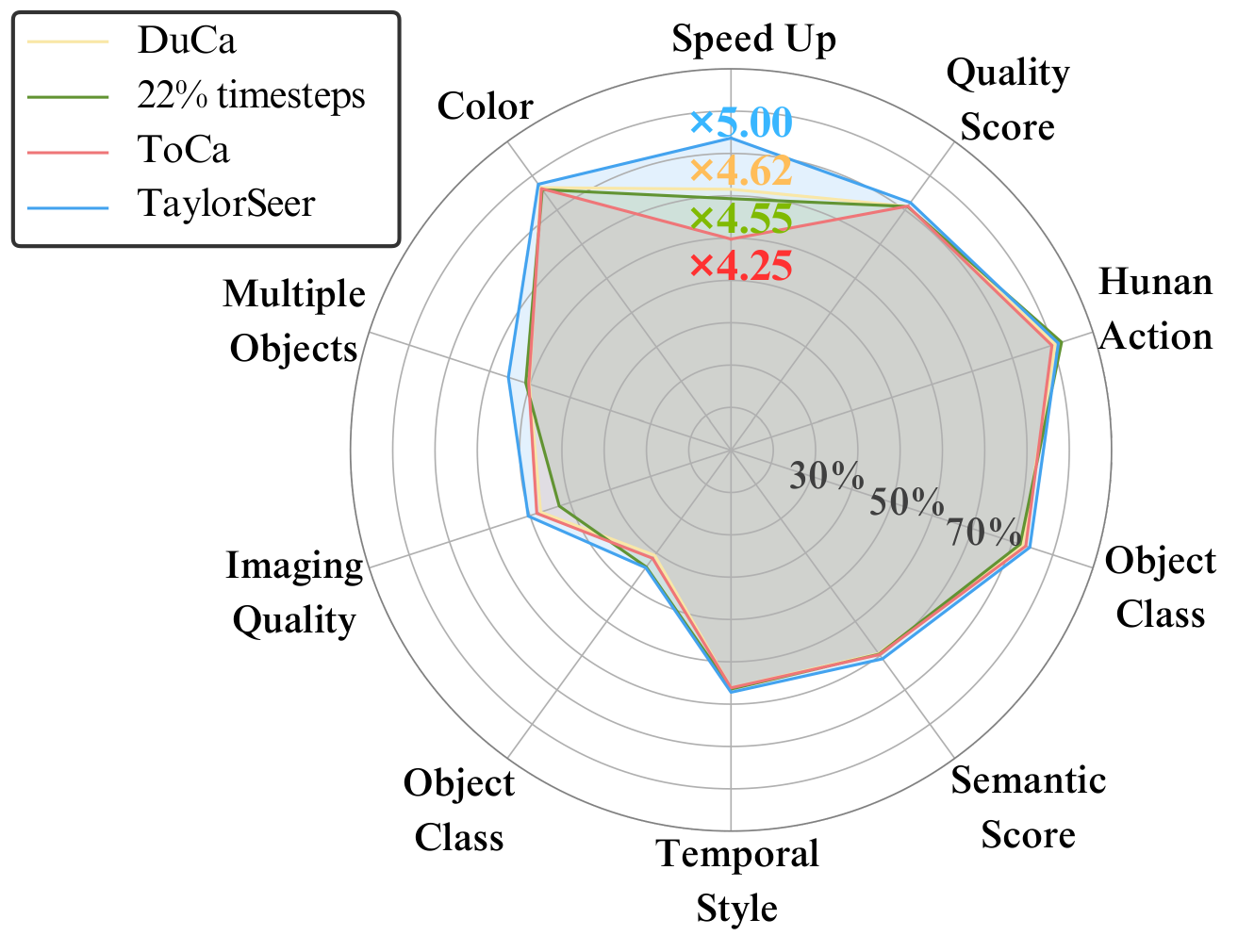}
\vspace{-4mm}
\caption{VBench metrics and acceleration ratios of  \textit{TaylorSeer}.}
\label{fig:radar}
\vspace{-5mm}
\end{minipage}
\hfill
\begin{minipage}{0.71\linewidth}
\centering
\includegraphics[width=0.98\linewidth]{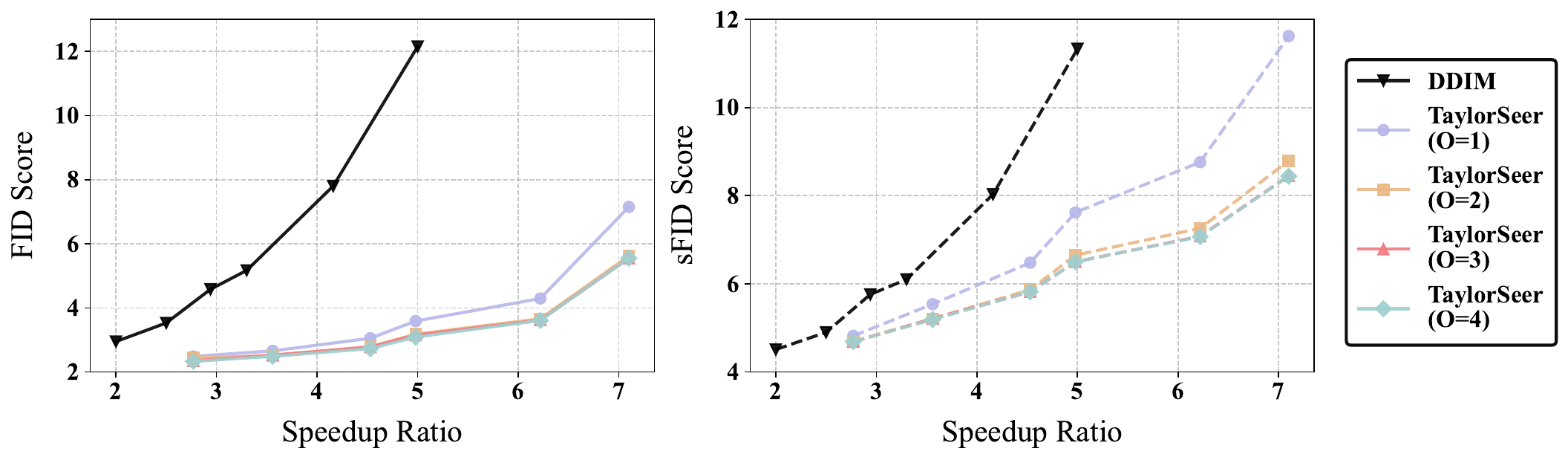}
\vspace{-4mm}
\caption{Comparison of 1st–4th order Taylor expansions for caching-based feature prediction. Higher-order expansions exhibit similar convergence and consistently outperform the 1st-order expansion at lower FLOPs, demonstrating the advantages of high-order approximation.}
\label{fig:flops-ablation}
\end{minipage}
\vspace{-4.5mm}
\end{figure*}

\vspace{-2mm}
\paragraph{Qualitative Study.}
\vspace{-3mm}

The qualitative results highlight \textit{TaylorSeer}'s ability to preserve video quality while accelerating computation. In the \textit{``a fire hydrant and a stop sign"} scenario, \textit{TaylorSeer} accurately generates the correct text, while methods like HunyuanVideo introduce a spelling error in \textit{``STOP"}, affecting semantic accuracy. In the \textit{``a donut and a suitcase"} case, \textit{TaylorSeer} maintains exceptional visual consistency, whereas other methods fail to generate the suitcase entirely. In the \textit{``a motorcycle turning a corner"} case, \textit{TaylorSeer} preserves motion blur details and trajectory smoothness, leading to more realistic video. Across these diverse scenarios, \textit{TaylorSeer} consistently outperforms other methods in fidelity and quality

\vspace{-1mm}
\subsection{Class-Conditional Image Generation}  
\vspace{-1mm}

We compared \textit{TaylorSeer} with methods such as ToCa, FORA, DuCa, and reduced DDIM steps on DiT-XL/2~\cite{DiT}, demonstrating that \textit{TaylorSeer} significantly outperforms the others in both acceleration ratio and generation quality. \textit{TaylorSeer}($\mathcal{N}$=3, $\mathcal{O}$=3) achieves an FID-50k of 2.34 while providing a \textbf{2.77$\times$ acceleration}, showing that higher-order Taylor expansions preserve feature quality at high acceleration ratios without incurring additional computational overhead. At a \textbf{4.53$\times$ acceleration}, our method ($\mathcal{N}$=5, $\mathcal{O}$=3) maintains an FID of 2.65, outperforming state-of-the-art models such as ToCa and DuCa. Notably, as the acceleration ratio increases beyond 3.5$\times$, other methods like FORA, ToCa, and DuCa exhibit significant degradation in FID, leading to severe deterioration in image quality, while \textit{TaylorSeer} consistently maintains superior performance. This robustness under extreme acceleration conditions further validates the effectiveness of our approach in predicting future features, which minimizes error accumulation compared to cache-then-reuse strategies that struggle to adapt to higher acceleration demands.
\vspace{-1mm}

\subsection{Ablation Studies}
\vspace{-2mm}
We conduct ablation experiments on DiT-XL/2, focusing on the impact of the interval parameter $\mathcal{N}$ and the Taylor expansion order $\mathcal{O}$ on computational efficiency and generation quality.
Results indicate that when $\mathcal{O}$=0 (direct feature reuse), performance degrades rapidly as $\mathcal{N}$ increases. Introducing higher-order Taylor expansions significantly improves generation quality, especially in long-interval scenarios. Even a first-order approximation ($\mathcal{O}$=1) substantially enhances performance, reducing FID from non-generation to 4.29 at $\mathcal{N}$=7. Higher-order expansions ($\mathcal{O} \geq 2$) further improve generation quality, capturing nonlinear feature variations effectively. In high-acceleration settings ($\mathcal{N}$=5 or $6$), third-order expansion maintains low FID (2.78–3.15), outperforming direct reuse. While $\mathcal{O} \geq 3$ continues to refine results, improvements saturate beyond the third order, as observed in both sFID and FID trends. 
Overall, \textit{TaylorSeer} demonstrates that Taylor expansion-based forecasting effectively balances efficiency and quality, particularly in high-acceleration scenarios. Higher-order expansions ($\mathcal{O} \geq 3$) enable high-quality generation with reduced computational cost, making the method suitable for real-time or resource-constrained applications.
\textit{Detailed results are provided in the supplementary materials.}
\vspace{-8mm}

%% file: tab/FLUX-Metrics.tex
\begin{table*}[htbp]
\centering
\caption{\textbf{Quantitative comparison in text-to-image generation} for FLUX on Image Reward.
}
\vspace{-3mm}
\setlength\tabcolsep{7.0pt} 
%\belowrulesep=0pt
%\aboverulesep=0pt
  \small
  \resizebox{0.98\textwidth}{!}{
  \begin{tabular}{l | c | c  c | c  c | c | c |c |c|c}
    \toprule
    {\bf Method} & {\bf Efficient} &\multicolumn{4}{c|}{\bf Acceleration} &{\bf Image Reward $\uparrow$} &\bf CLIP$\uparrow$ & \multirow{2}{*}{\bf PSNR$\uparrow$} & \multirow{2}{*}{\bf SSIM$\uparrow$} & \multirow{2}{*}{\bf LPIPS$\downarrow$}\\
    \cline{3-6}
    {\bf FLUX.1\citep{flux2024}} & {\bf Attention \citep{Dao2022FlashAttentionFA}} & {\bf Latency(s) $\downarrow$} & {\bf Speed $\uparrow$} & {\bf FLOPs(T) $\downarrow$}  & {\bf Speed $\uparrow$} & \bf DrawBench &\bf Score & & & \\
    \midrule
  
  $\textbf{[dev]: 50 steps}$ 
  %\citep{flux2024}         
                           & \ding{52}  &  {17.20}  & {1.00$\times$} & {3719.50}   & {1.00$\times$} & {0.9898}  &{19.604}    & - &  -  & -\\

  {$60\%$\textbf{ steps}}  & \ding{52}  &  {10.49}  & {1.64$\times$} & {2231.70}   & {1.67$\times$} & {0.9739}  &{19.526}     & 30.310 & 0.7819  & 0.2461\\

  {$50\%$\textbf{ steps}}  {\textcolor{red}{†}}  & \ding{52}  &  {8.80}   & {1.95$\times$} & {1859.75}  & {2.00$\times$} & {0.9429}  &{19.325}          & 29.576 &  0.7337 & 0.3106 \\
  
  {$40\%$\textbf{ steps}} {\textcolor{red}{†}}  & \ding{52}  &  {7.11}   & {2.42$\times$} & {1487.80}   & {2.62$\times$} & {0.9317}  &{19.027}           & 29.122 & 0.6971 & 0.3619 \\
  
  {$34\%$\textbf{ steps}} {\textcolor{red}{†}}  & \ding{52}  &  {6.09}   & {2.82$\times$} & {1264.63}   & {3.13$\times$} & {0.9346}  &{18.904}           & 28.881 &0.6776   & 0.3913\\

  \midrule

  {$\Delta$-DiT} ($\mathcal{N}=2$) {\textcolor{red}{†}} & \ding{52}  &  {11.87}  & {1.45$\times$} & {2480.01}   & {1.50$\times$} & {0.9316}  &{19.350}    & 29.399 &  0.7227 & 0.3301\\
  
  {$\Delta$-DiT} ($\mathcal{N}=3$) {\textcolor{red}{†}}  & \ding{52}  &  {8.81}  & {$1.95\times$} & {1686.76}   & {2.21$\times$} & {0.8561}  &{18.833}      & 28.794 & 0.6665 &0.4133\\
  %\midrule
  
  $\textbf{FORA}$ $(\mathcal{N}=3)$ {\textcolor{red}{†}}
  \citep{selvaraju2024fora}& \ding{52}  &  {7.08}   & {2.43$\times$} & {1320.07}   & {2.82$\times$} & {0.9227}    &{18.950}   & 30.652 & 0.7666 & 0.2450\\

  \textbf{TeaCache} $({l}=0.4)$ {\textcolor{red}{†}}  \citep{liu2024timestep}& \ding{52}  & 7.56 & $ 2.28\times$ & 1412.90 & 2.63$\times$  &  0.9687  & 19.096& 28.154 & 0.4674 &0.6624\\ 
  
  \rowcolor{gray!20}
  
  $\textbf{TaylorSeer} $ $(\mathcal{N}=3,O=2)$ %ATA
                           & \ding{52}  &  {6.59}   & {2.61$\times$} &  \textbf{1320.07}   & \textbf{2.82$\times$} &   \textbf{1.0181} & \textbf{19.397} & \textbf{30.762} & \textbf{0.7818} & \textbf{0.2300}\\
    
    \midrule
  % $\textbf{FORA}$ $(\mathcal{N}=4)$ {\textcolor{red}{†}} 
  % \citep{selvaraju2024fora}& \ding{52}  &  {5.43}   & {3.17$\times$} & {967.91}   & {3.84$\times$} & {0.8675}    &{18.560} & 29.975 &  0.7217& 0.3007\\

  $\textbf{FORA}$ $(\mathcal{N}=5)$ {\textcolor{red}{†}} 
  \citep{selvaraju2024fora}& \ding{52}  &  {5.17}   & {3.33$\times$} & {893.54}   & {4.16$\times$} & {0.8235}    &{18.279} & 28.430 &  0.6030& 0.4915\\
    
  $\textbf{\texttt{ToCa}}$ $(\mathcal{N}=5)$%$(\mathcal{N}=5,R=90\%)$
  \citep{zou2024accelerating} 
                           & \ding{56}  &  {10.80}   & {1.59$\times$} & {1126.76}   & {3.30$\times$} & {0.9731}   &{19.030}  & {29.642} & 0.7082 & 0.3319\\

  $\textbf{\texttt{DuCa}} (\mathcal{N}=5)$      %$(\mathcal{N}=5,R=90\%)$ %ATA
  \citep{zou2024DuCa}
                           & \ding{52}  &  {5.88}    & {2.93$\times$} & {1078.34}   & {3.45$\times$} &{0.9896}   &{\textbf{19.595}}  & 29.413 & 0.7142 & 0.3082\\
                           
  \textbf{TeaCache} $({l}=0.6)$ {\textcolor{red}{†}}  \citep{liu2024timestep}& \ding{52}  & 6.07 & 2.83$\times$ & 1115.44 & 3.33$\times$  &  0.9670  & 18.783& 28.147 & 0.4674 &0.6624\\ 

  \rowcolor{gray!20}
  $\textbf{TaylorSeer} $ $(\mathcal{N}=4,O=2)$ %ATA
                           & \ding{52}  &  5.61   &  3.07$\times$ & {967.91}   & {3.84$\times$}& 1.0204  & 19.433 & \textbf{29.863} & \textbf{0.7273} & \textbf{0.2993}\\

  \rowcolor{gray!20}
  $\textbf{TaylorSeer} $ $(\mathcal{N}=5,O=2)$ %ATA
                           & \ding{52}  &  {5.82}    &  {2.96}{$\times$} &  {\textbf{893.54}}  &  {\textbf{4.16}$\times$} &\textbf{1.0296}   &{19.437} & 29.330 &  0.6994 & 0.3456\\
  \midrule
  
  $\textbf{FORA}$ $(\mathcal{N}=6)$ {\textcolor{red}{†}} 
  \citep{selvaraju2024fora}& \ding{52}  &  {4.53}   & {3.80$\times$} & {744.81}   & {4.99$\times$} & {0.7761}    &{17.986} & 28.360 & 0.6001 & 0.5177\\

  $\textbf{\texttt{ToCa}}$ $(\mathcal{N}=8)$ {\textcolor{red}{†}}  %$(\mathcal{N}=8,R=90\%)$
  \citep{zou2024accelerating} 
                           & \ding{56}  &  {8.47}   & {2.03$\times$} & {784.54}   & {4.74$\times$} & {0.9086}   &{18.380} & \textbf{28.961} & 0.6248 & 0.4470\\
  $\textbf{\texttt{DuCa}}$ $(\mathcal{N}=6)$ {\textcolor{red}{†}}  %$(\mathcal{N}=6,R=90\%)$ %ATA
  \citep{zou2024DuCa}
                           & \ding{52}  &  {4.89}    & {3.52$\times$} & {816.55}   & {4.56$\times$} &{0.9470}   &{19.082} &  28.672&  0.6228& 0.4182\\                         
  \textbf{TeaCache} $({l}=0.8)$ {\textcolor{red}{†}}  \citep{liu2024timestep}& \ding{52}  & 4.98 & 3.58$\times$ & 892.35 & 4.17$\times$  &  0.8683  & 18.500 & 28.146 &  0.4692& 0.6640\\ 
  
  \rowcolor{gray!20}
  $\textbf{TaylorSeer} $ $(\mathcal{N}=6,O=1)$ %ATA
                           & \ding{52}  &  4.87    & 3.53$\times$ & \textbf{744.81}  & \textbf{4.99$\times$} &{0.9953}   &\textbf{19.637} & 28.826 & 0.6544 & 0.4097\\
  \rowcolor{gray!20}
  $\textbf{TaylorSeer} $ $(\mathcal{N}=6,O=2)$ %ATA
                           & \ding{52}  &  {5.19}  & {3.31$\times$}  & {744.81}   & {4.99$\times$} & \bf{1.0039}   &{19.427} & 28.945 & \textbf{0.6556}& \textbf{0.4020}\\

    \bottomrule
  \end{tabular}}
  
  \label{table:FLUX-Metrics}
\footnotesize
\begin{itemize}\item \textcolor{red}{†} Methods exhibit significant degradation in Image Reward, leading to severe deterioration in image quality.\end{itemize}
%\vspace{-3mm}
\end{table*}

%% file: tab/HunyuanVideo-Metrics.tex
\begin{table*}[htbp]
\centering
\caption{\textbf{Quantitative comparison in text-to-video generation} for HunyuanVideo on VBench.
}
\vspace{-3mm}
\setlength\tabcolsep{5.0pt} 
%\belowrulesep=0pt
%\aboverulesep=0pt
  \small
  \resizebox{0.98\textwidth}{!}{
  \begin{tabular}{l | c | c  c | c  c | c | c c c}
    \toprule
    {\bf Method} & {\bf Efficient} &\multicolumn{4}{c|}{\bf Acceleration} &{\bf VBench $\uparrow$} & \multicolumn{3}{c}{\bf Quality Metrics} \\
    \cline{3-6}
    \cline{8-10}
    {\bf HunyuanVideo\citep{sun_hunyuan-large_2024}} & {\bf Attention \citep{Dao2022FlashAttentionFA}} & {\bf Latency(s) $\downarrow$} & {\bf Speed $\uparrow$} & {\bf FLOPs(T) $\downarrow$}  & {\bf Speed $\uparrow$} & \bf Score(\%) & \bf PSNR$\uparrow$ & \bf SSIM$\uparrow$ & \bf LPIPS$\downarrow$ \\
    \midrule
  
  $\textbf{Original: 50 steps}$ 
                           & \ding{52}  &  {318.24}  & {1.00$\times$} & {29773.0}   & {1.00$\times$} & {80.66} & - & - & -      \\ 
  {$22\%$\textbf{ steps}}  & \ding{52}  &  {70.34}   & {4.55$\times$} & {6550.1}   & {4.55$\times$} & {78.74} & 16.816 & 0.5708 & 0.3990            \\
                           
  \midrule
$\textbf{AdaCache}$ \citep{kahatapitiyaAdaptiveCachingFaster2024}& \ding{52}  &  {120.59}   & {2.64$\times$} & {11182.5} & {2.66$\times$} & {80.25} & {17.089} & {0.6958} & {0.3197}  \\
$\textbf{FORA}(N=3)$ \citep{selvaraju2024fora}& \ding{52}  &  {112.57}   & {2.83$\times$}  &9929.2 & 3.00$\times$ & 80.26 & 19.447 & 0.7688 & 0.1960 \\
\textbf{TeaCache}$({l}=0.2)$ \cite{liu2024timestep} & \ding{52}  &  {109.70}   & {2.90$\times$} & 10023.9 & 2.97$\times$ & 80.44 & 20.230 & 0.7927 & 0.1759 \\
\rowcolor{gray!20}
$\textbf{{TaylorSeer}}(N=3, O=1)$& \ding{52}  &  {113.44}   & {2.81$\times$}  & \textbf{9929.2} & \textbf{3.00$\times$} & \textbf{80.74} & \textbf{21.423} & \textbf{0.8335} & \textbf{0.1353} \\
    
  \midrule

  $\textbf{FORA}(N=5)$ 
  \citep{selvaraju2024fora}& \ding{52}  &  {67.19}   & {4.74$\times$} & {5960.4}   & {5.00$\times$} & {78.83} & 16.072 & 0.6334 & 0.3457     \\
  $\textbf{\texttt{ToCa}}$ $(\mathcal{N}=5,R=90\%)$
  \citep{zou2024accelerating} 
                           & \ding{56}  &  {77.82*}   & {4.09$\times$} & {7006.2}   & {4.25$\times$} & {78.86} & 16.143 & 0.6424 & 0.3432    \\
  $\textbf{\texttt{DuCa}} $ $(\mathcal{N}=5,R=90\%)$ 
  \citep{zou2024DuCa}
                           & \ding{52}  &  {71.10 }   & {4.58$\times$} & {6483.2}   & {4.48$\times$} & {78.72} & 16.169 & 0.6443 & 0.3516    \\
                           
\textbf{TeaCache}$({l}=0.4)$ \cite{liu2024timestep} & \ding{52} & 67.36 & $4.72\times$ & 6550.1 & 4.55$\times$ & 79.36 & 16.072 & 0.6216 & 0.4377 \\
  \rowcolor{gray!20}
  $\textbf{TaylorSeer} $ $(\mathcal{N}=5,O=1)$ %ATA
                           & \ding{52}  &  {68.42 }    & {4.65$\times$} & {5960.4}  & {5.00$\times$} &\bf{79.93} & \textbf{16.796} & 0.7039 & \textbf{0.2691}   \\
  \rowcolor{gray!20}
  $\textbf{TaylorSeer} $ $(\mathcal{N}=6,O=1)$ %ATA
                           & \ding{52}  &  \textbf{61.99}    & \textbf{5.13$\times$} & \textbf{5359.1}  & \textbf{5.56$\times$} &{79.78} & {16.241} & \textbf{0.7215} & {0.3009}   \\
    \bottomrule
  \end{tabular}}
  
  \label{table:HunyuanVideo-Metrics}
\vspace{-3mm}
\end{table*}

%% file: tab/DiT-Metrics.tex
\begin{table*}[htbp]
\centering
\vspace{-0.3cm}
\caption{\textbf{Quantitative comparison on class-to-image generation} on ImageNet with \text{DiT-XL/2.}}
\vspace{-0.2cm}
\setlength\tabcolsep{7pt} 
\small
\renewcommand{\arraystretch}{0.8}
\begin{tabular}{l | c | c c c | c c |c}
\toprule
\bf Method  & \makecell{\bf Efficient \\ \bf Attention} & \bf Latency(s) $\downarrow$ & \bf FLOPs(T) $\downarrow$ & \bf Speed $\uparrow$  & \bf FID $\downarrow$ & \bf sFID $\downarrow$ &   \makecell{\bf Inception\\ \bf Score} $\uparrow$  \\
\midrule
{\textbf{$\text{DDIM-50 steps}$}} & \ding{52}& {0.428}  & {23.74}  & {1.00$\times$}  &  {{2.32}} &  {{4.32}}   &{241.25}\\
{\textbf{$\text{DDIM-25 steps}$}} & \ding{52}& {0.230}  & {11.87}  & {2.00$\times$}  &  {{3.18}} &  {{4.74}}   &{232.01}\\
{\textbf{$\text{$\Delta$-DiT}$}}($\mathcal{N}=2$)  & \ding{52}& {0.246}  & {18.04}  & {1.31$\times$}  &  {{2.69}} &  {{4.67}} &{225.99}\\
{\textbf{$\text{$\Delta$-DiT}$}}($\mathcal{N}=3$) {\textcolor{red}{†}}  & \ding{52}& {0.173}  & {16.14}  & {1.47$\times$}  &  {3.75} &  {5.70} &{207.57}\\

\midrule     

%2倍
{\textbf{$\text{DDIM-20 steps}$}} & \ding{52}& {0.191}  & {9.49}  & {2.50$\times$}  &  {{3.81}} &  {{5.15}} &{221.43}\\
\textbf{FORA} ($\mathcal{N}=3$) & \ding{52} & 0.197 & 8.58 & {2.77$\times$}  & 3.55 & 6.36  &{229.02}\\
\textbf{\texttt{ToCa}} ($\mathcal{N}=3$)  & \ding{56}& {0.216} & {10.23} & {2.32$\times$} & {2.87}   & {4.76} &{235.21}\\
\textbf{\texttt{DuCa}} ($\mathcal{N}=3$) & \ding{52}& {0.208} & {9.58} & {2.48$\times$} & {2.88}  & \textbf{4.66}  &{233.37}\\
\rowcolor{gray!20}
$\textbf{TaylorSeer} $ $(\mathcal{N}=3,O=3)$& \ding{52}& {0.248} & {\textbf{8.56}} & {\textbf{2.77}$\times$} & {\textbf{2.34}}  & {4.69} &{\textbf{238.42}} \\
%3.5倍
\midrule 
{\textbf{$\text{DDIM-15 steps}$}} {\textcolor{red}{†}} & \ding{52}& {0.152}  & {7.12}  & {3.33$\times$}  &  {{5.17}} &  {{6.11}} &{206.33}\\
\textbf{FORA} ($\mathcal{N}=4$) {\textcolor{red}{†}}& \ding{52} & 0.169 & 6.66 & {3.56$\times$}  & 4.75 & 8.43  &{213.72}\\
\textbf{\texttt{ToCa}} ($\mathcal{N}=5$)  \textcolor{red}{†} & \ding{56}& {0.173} & {6.77} & {3.51$\times$} & {6.20}   & {7.17} &{200.04}\\

\textbf{\texttt{DuCa}} ($\mathcal{N}=4$){\textcolor{red}{†}}  & \ding{52}& {0.144} & {7.61} & {3.11$\times$} & {3.42}  & \textbf{4.94}  &{225.19}\\
\rowcolor{gray!20}
$\textbf{TaylorSeer} $ $(\mathcal{N}=4,O=4)$ & \ding{52}& {0.220} & {\textbf{6.66}} & {\textbf{3.56}$\times$} & {\textbf{2.49}}  & {5.19}  &{\textbf{235.83}}\\
\midrule 

%大于3.5倍

{\textbf{$\text{DDIM-12 steps}$}} {\textcolor{red}{†}} & \ding{52}& {0.128}  & {5.70}  & {4.17$\times$}  &  {{7.80}} &  {{8.03}} &{184.50}\\
{\textbf{$\text{DDIM-10 steps}$}} {\textcolor{red}{†}} & \ding{52}& {0.115}  & {4.75}  & {5.00$\times$}  &  {{12.15}} &  {{11.33}} &{159.13}\\
\textbf{FORA} ($\mathcal{N}=5$) {\textcolor{red}{†}}& \ding{52} & 0.149 & 5.24 & {4.53$\times$}  & 6.58 & 11.29  &{193.01}\\

\textbf{\texttt{ToCa}} ($\mathcal{N}=6$)  \textcolor{red}{†} & \ding{56}& {0.163} & {6.34} & {3.75$\times$} & {6.55}   & {7.10} &{189.53}\\

\textbf{\texttt{DuCa}} ($\mathcal{N}=5$){\textcolor{red}{†}}  & \ding{52}& {0.154} & {6.27} & {3.78$\times$} & {6.06}  & 6.72  &{198.46}\\

\rowcolor{gray!20}
$\textbf{TaylorSeer} $ $(\mathcal{N}=5,O=3)$& \ding{52}& {0.180} & {\textbf{5.24}} & {\textbf{4.53}$\times$} & {\textbf{2.65}}  & {\textbf{5.36}}  &{\textbf{231.59}}\\

\rowcolor{gray!20}
\bottomrule
\end{tabular}
\label{table:DiT_Metrics}
\vspace{0mm}
%\begin{tablenotes}
%\footnotesize
%\item \textcolor{red}{†} Methods exhibit significant degradation in FID when the acceleration ratio exceeds 3.5×, leading %to severe deterioration in image quality.
%\end{tablenotes}
\footnotesize
\begin{itemize}\item \textcolor{red}{†} Methods exhibit significant degradation in FID, leading to severe deterioration in image quality.\end{itemize}
\vspace{-5mm}
\end{table*}

%% file: sec/5_conclusion.tex
%\vspace{-1mm}
\section{Conclusion}
\vspace{-3mm}

Traditional feature caching methods follow the paradigm of \textit{``cache-then-reuse''}, which directly reuses the features stored at previous timesteps in the following timesteps, and suffer
from serve drop in generation quality at high acceleration ratios.
In this paper, motivated by the surprisingly stable trajectory of features at different timesteps, we propose \textit{``cache-then-forecast''}, which formulates feature caching as a sequential prediction problem and solves it with Taylor expansion. 
The high-order Taylor expansions in our method can capture complex feature trajectories more rigorously, making it maintain the generation quality in high-acceleration ratios, where all previous caching methods totally fail. Experiments across architectures including  DiT, FLUX, and HunyuanVideo demonstrate significant acceleration (4.53$\times$–5.56$\times$) without quality degradation. 
We hope \textit{TaylorSeer} can move the paradigm of feature caching methods from reusing to forecasting.

\vspace{-3mm}
\section*{Acknowledgement}
\vspace{-3mm}
This work was partially supported by the Dream Set Off - Kunpeng\&Ascend Seed Program.

%% file: sec/X_suppl.tex
\clearpage
\setcounter{page}{1}
\maketitlesupplementary

\section{Experimental Details}
\label{sec:Experimental Details}
In this section, more details of the experiments are provided.

\subsection{Model Configuration}
As mentioned in 4.1.1 , experiments on three models from different tasks, FLUX~\cite{flux2024} for text-to-image generation, HunyuanVideo~\cite{li_hunyuan-dit_2024,sun_hunyuan-large_2024} for text-to-video generation, and DiT~\cite{DiT} for class-conditional image generation, are presented. In this section, a more detailed hyperparameter configuration scheme is provided.
\begin{enumerate}
    \item[$\bullet$] \textbf{FLUX}: The FORA~\cite{selvaraju2024fora} method employs a uniform activation interval with $\mathcal{N}$=3. The ToCa~\cite{zou2024accelerating} method uses $\mathcal{N}$=4 with a caching ratio of 90\%, adopting a non-uniform activation interval, with sparse activations at the beginning and dense activations towards the end, utilizing an attention-based token selection method. The DuCa~\cite{zou2024DuCa} method sets conservative caching steps on even-numbered steps following fresh steps, while aggressive caching steps are set on odd-numbered steps. The activation interval and caching ratio are consistent with the ToCa method, also using a non-uniform activation interval and employing the attention-based token selection method.

    \item[$\bullet$] \textbf{HunyuanVideo}: The FORA~\cite{selvaraju2024fora} method utilizes an activation interval of $\mathcal{N}$=5, whereas both the ToCa~\cite{zou2024accelerating} and DuCa methods employ $\mathcal{N}$=4 with a caching ratio of 90\%. For each activation step (complete computational step), aggressive caching is applied to odd-numbered steps, and conservative caching is applied to even-numbered steps. Due to memory limitations that result in an "out of memory" error when using a non-uniform activation scheme, the ToCa method in HunyuanVideo is configured with a uniform activation interval, as indicated by a $^*$ in the corresponding table.

    \item[$\bullet$] \textbf{DiT}: The FORA~\cite{selvaraju2024fora} method uses a uniform activation interval with $\mathcal{N}$=3. The ToCa~\cite{zou2024accelerating} method also uses $\mathcal{N}$=3 with an average caching ratio of $R = 95\%$, employing a non-uniform activation interval, with sparse activations at the beginning and dense activations towards the end, utilizing the attention-based token selection method. The DuCa~\cite{zou2024DuCa} method sets conservative caching steps on even-numbered steps following fresh steps, while aggressive caching steps are set on odd-numbered steps. The activation interval and caching ratio are consistent with the ToCa method, also using a non-uniform activation interval with sparse-to-dense activation and employing the attention-based token selection method. $\Delta$-DiT adopts a layer-skipping strategy where, in the early stages (49-25 steps), layers 14-27 are skipped, and in the later stages (24-0 steps), layers 0-13 are skipped.
\end{enumerate}

\section{Supplementary Results for Ablation Studies}
We conduct ablation experival parameter $\mathcal{N}$ and the Taylor expansion order $\mathcal{O}$ on computatments on DiT-XL/2~\cite{DiT} to evaluate \textit{TaylorSeer}, focusing on the impact of the interional efficiency and generation quality. The results demonstrate the importance of these design choices in balancing performance and speed.
\vspace{-3mm}
\input{tab/DiT-Metrics-Ablation}

\section{Anonymous Page for Video Presentation}

To further showcase the advantages of TaylorSeer in video generation, we have created an anonymous GitHub page. For a more detailed demonstration, please visit \url{https://taylorseer.github.io/TaylorSeer/}. Additionally, the videos are also available in the Supplementary Material.

\section{Supplementary Visualization Examples}
To further illustrate the qualitative improvements of our method, we present visualization examples on FLUX and HunyuanVideo. These results showcase the superior fidelity and consistency of our method in generating high-quality outputs across diverse scenarios.

\section{Supplementary Visualization of Feature Trajectories in Diffusion Models}
In this section, we provide additional visualizations of feature trajectories and their derivatives in diffusion models. These results further illustrate the stability and predictability of feature dynamics across different timesteps, supporting our findings in the main text. The PCA projections of features (0th-order) and their derivatives (1st to 4th-order) demonstrate consistent patterns, highlighting the potential for efficient feature prediction in diffusion models.

\begin{figure}[h!]
    \centering
    \includegraphics[width=\linewidth]{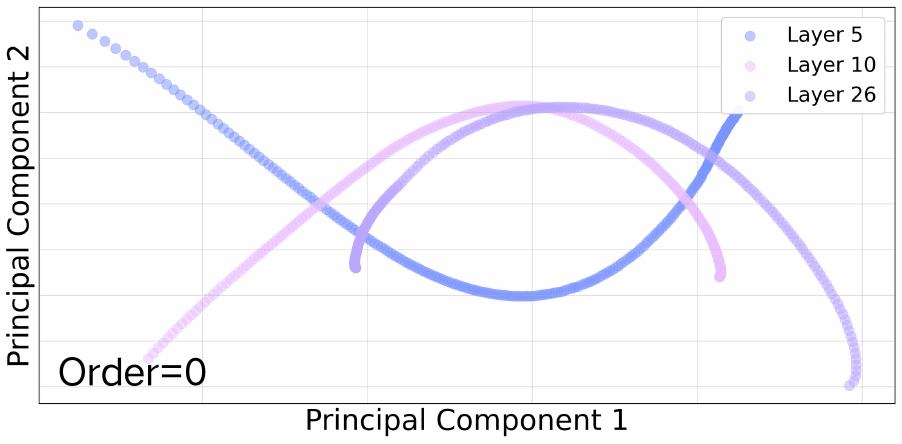}
    \caption{\textbf{PCA projections of features in diffusion models.} The features at different timesteps form stable trajectories, demonstrating the predictability of feature evolution over time.}
    \label{fig:DiT_Feature0}
\end{figure}

\begin{figure}[h!]
    \centering
    \includegraphics[width=\linewidth]{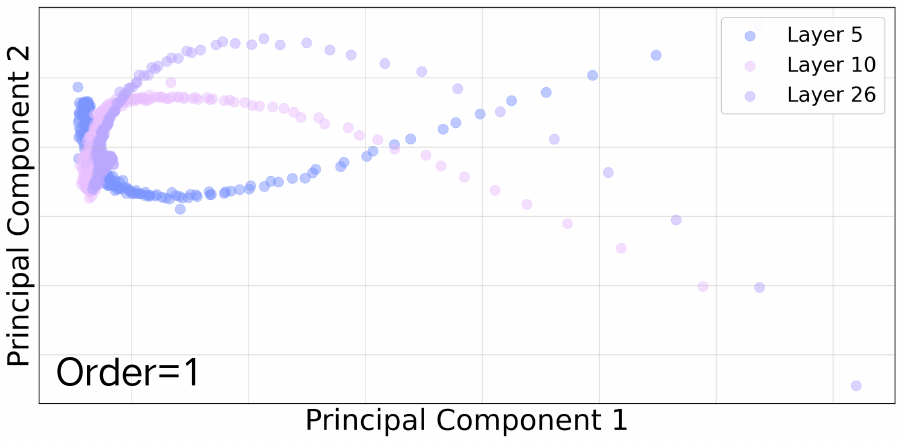}
    \caption{\textbf{PCA projections of first-order feature derivatives.} The first-order derivatives exhibit consistent patterns, further supporting the predictability of feature dynamics in diffusion models.}
    \label{fig:DiT_Feature1}
\end{figure}

\begin{figure}[h!]
    \centering
    \includegraphics[width=\linewidth]{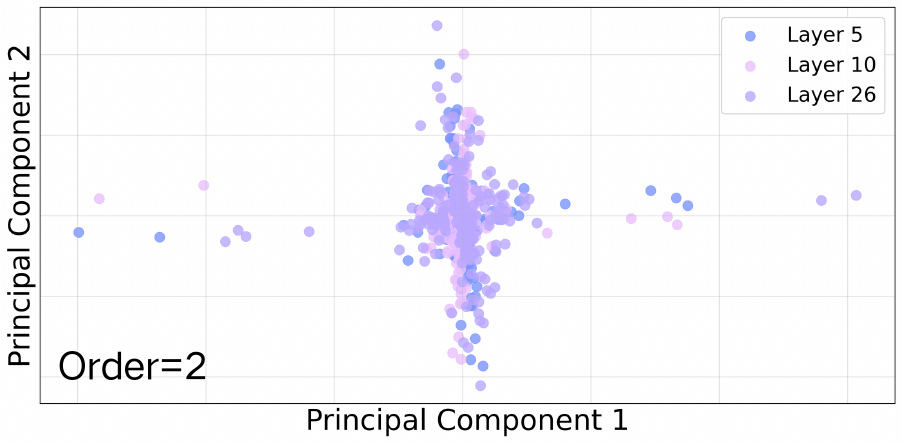}
    \caption{\textbf{PCA projections of second-order feature derivatives.} The second-order derivatives reveal higher-order dynamics, highlighting the smoothness of feature transitions.}
    \label{fig:DiT_Feature2}
\end{figure}

\begin{figure}[h!]
    \centering
    \includegraphics[width=\linewidth]{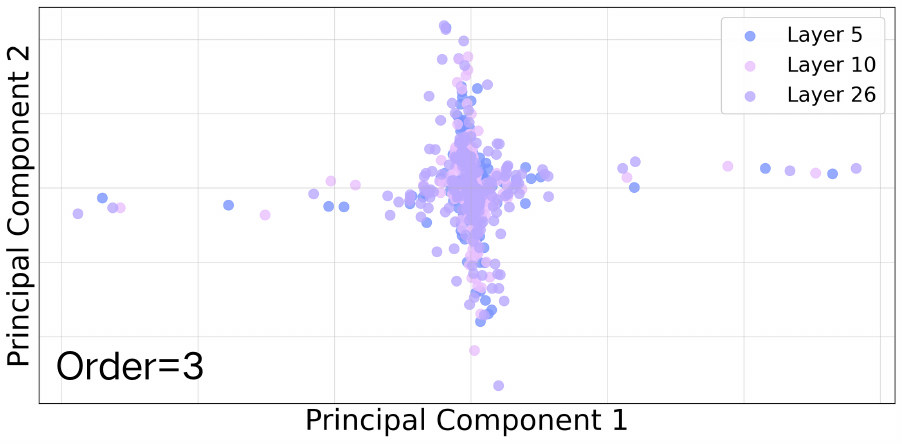}
    \caption{\textbf{PCA projections of third-order feature derivatives.} The third-order derivatives capture more complex temporal patterns, indicating the richness of feature dynamics.}
    \label{fig:DiT_Feature3}
\end{figure}

\begin{figure}[h!]
    \centering
    \includegraphics[width=\linewidth]{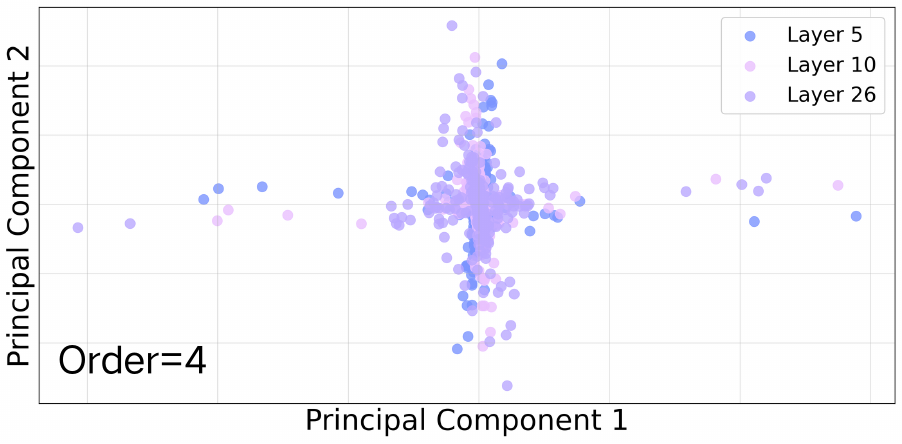}
    \caption{\textbf{PCA projections of fourth-order feature derivatives.} The fourth-order derivatives provide insights into fine-grained temporal variations, further validating the predictability of feature evolution.}
    \label{fig:DiT_Feature4}
\end{figure}

\begin{figure*}
    \centering
    \includegraphics[width=\linewidth]{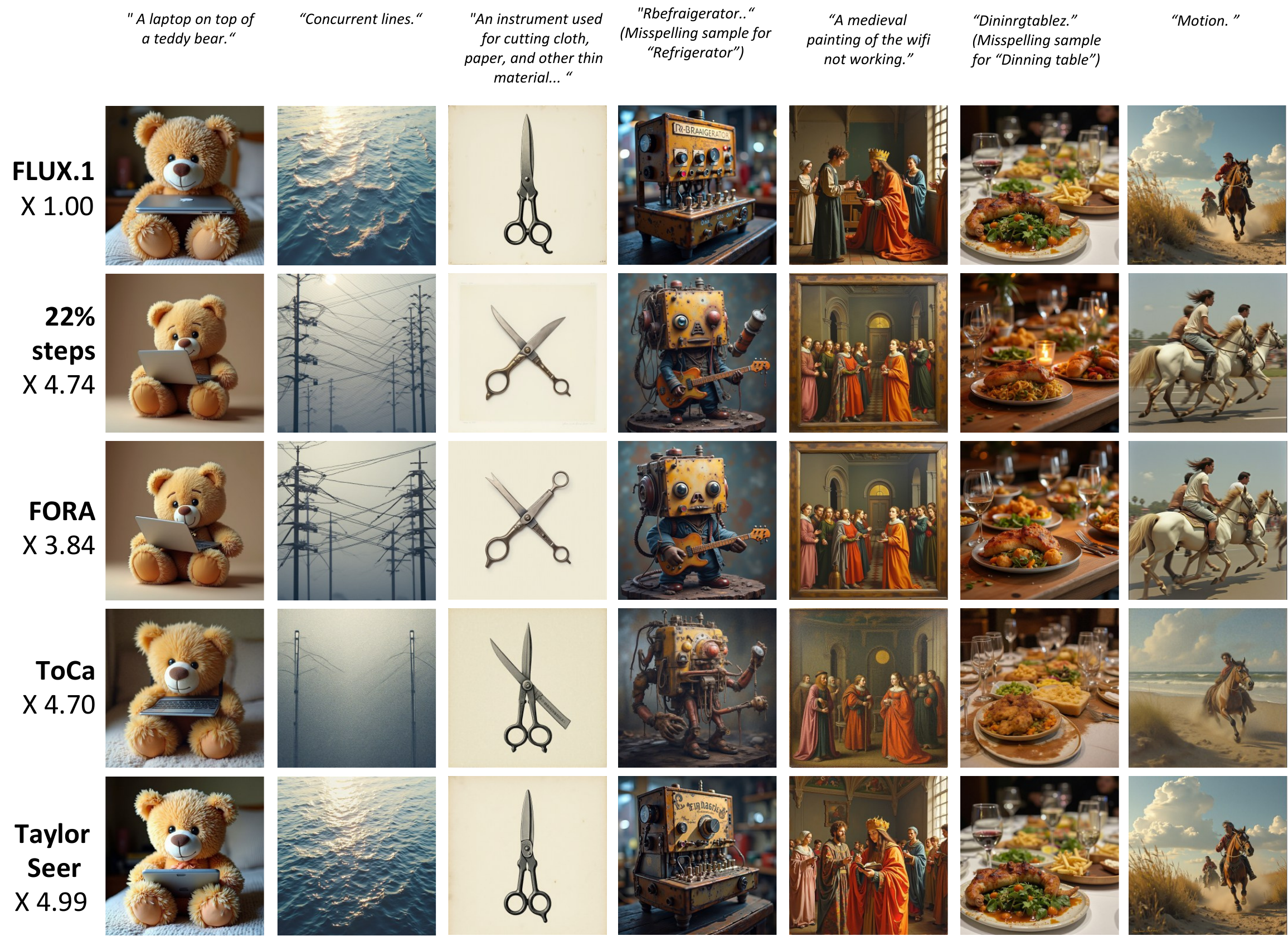}
    \caption{Visualization results for different acceleration methods on FLUX.1-dev.}
    \label{fig:Vis-FLUX}
\end{figure*}

\begin{figure*}
    \centering
    \includegraphics[width=\linewidth]{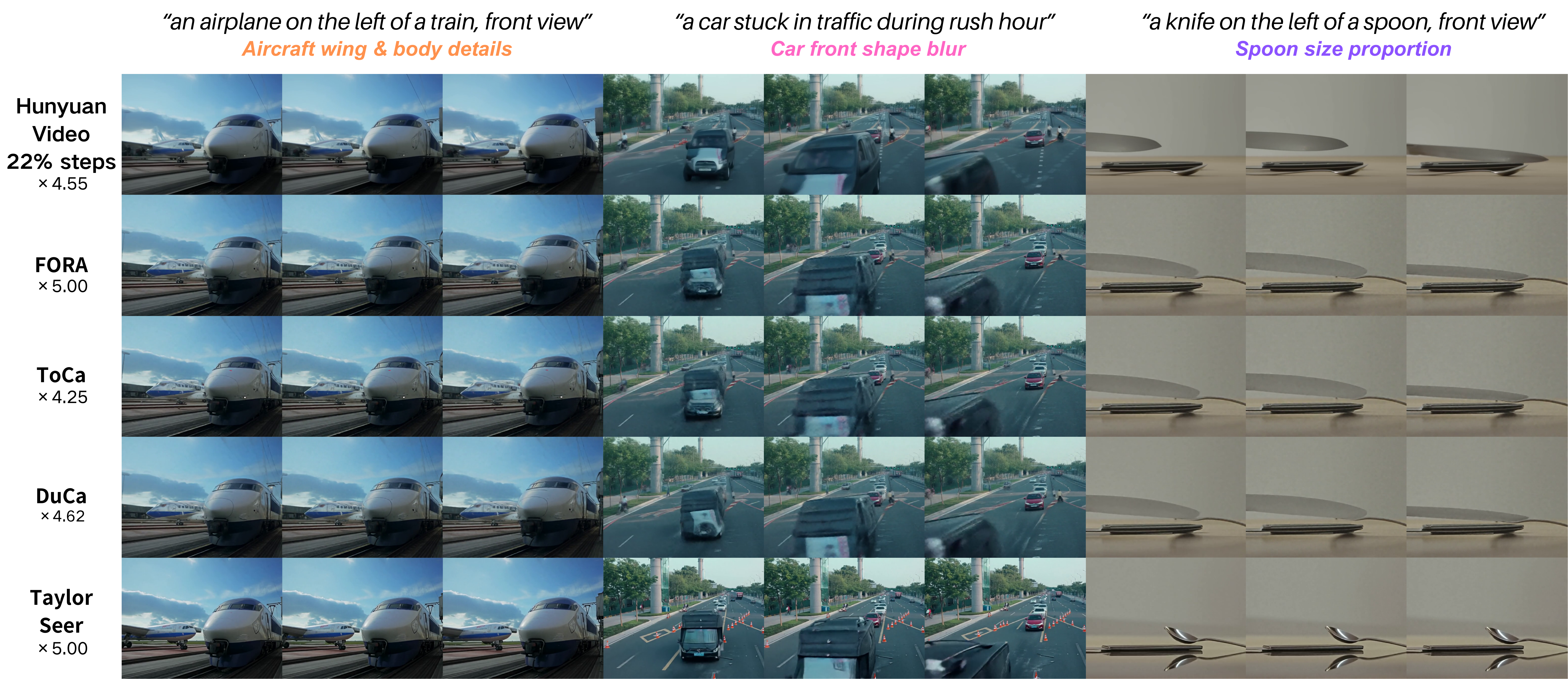}
    \caption{Visualization results for different acceleration methods on HunyuanVideo.}
    \label{fig:Vis-Huanyuan-2}
\end{figure*}

\section{More Results for TaylorSeer}

To further validate the broad applicability and adaptability of \textit{TaylorSeer}, we conducted comprehensive experiments on several other mainstream text-to-video DiT models, including HiDream~\cite{cai_hidream-i1_2025}, FramePack~\cite{zhang_packing_nodate}, and WAN2.1~\cite{wan_wan_2025}. The results demonstrate that \textit{TaylorSeer} consistently achieves superior performance across these platforms.

As a representative example, Table \ref{table:hidream-comparison-eng} presents a detailed performance comparison on the HiDream model. While achieving a 
4.0 $times$ acceleration, \textit{TaylorSeer} significantly outperforms the competing method, TeaCache, across all key evaluation metrics, including PSNR, SSIM, LPIPS, and ImageReward.

Furthermore, our project page showcases detailed comparison videos between \textit{TaylorSeer} and competing methods on the HunyuanVideo model, which visually demonstrate the significant advantages of our approach. Moreover, \textit{TaylorSeer} exhibits remarkable versatility by delivering unexpectedly impressive results when extended to \textit{super-resolution tasks} and audio generation.

\input{tab/HiDream-Metrics}

\section{Performance Analysis at Low Speedup Ratios}

Our study reveals that \textit{TaylorSeer} not only accelerates the inference process but can also surpass the baseline performance, particularly at reduced acceleration ratios. As detailed in Table \ref{table:FLUX-Metrics} and Table \ref{table:HunyuanVideo-Metrics}, our empirical results substantiate this finding. Specifically, when applied to FLUX, the ImageReward score improved from 0.9898 to 1.0181. Similarly, for HunyuanVideo, the VBench score increased from 80.66 to 80.74. These results validate the effectiveness of \textit{TaylorSeer} across the full spectrum of acceleration factors.

We attribute this counter-intuitive performance enhancement to the inherent parameter redundancy within the original large-scale models. This observation is analogous to established techniques such as model pruning and Low-Rank Adaptation (LoRA), where strategically reducing parameter utilization can sometimes lead to improved generalization and overall performance. We hypothesize that \textit{TaylorSeer} introduces a regularization-like effect during the inference steps. Since both FLUX and HunyuanVideo are based on Flow Matching, the Taylor approximation-based simplification at each step may guide the generation process along a more robust trajectory, effectively regularizing the model's output. This finding provides valuable insights for both model acceleration and future training methodologies, suggesting that targeted inference-time optimization can unlock latent performance gains.

%% file: tab/DiT-Metrics-Ablation.tex
\begin{table}[htbp]
\centering
\caption{\textbf{Ablation Study with Different Configurations} on ImageNet with \text{DiT-XL/2.}}
\label{table:Ablation}
\begin{tabularx}{\linewidth}{c|c  >{\centering\arraybackslash}c  |>{\centering\arraybackslash}X  >{\centering\arraybackslash}X }
\toprule
\textbf{Configuration} & \bf FLOPs(T)$\downarrow$ & \bf Speed$\uparrow$ &  \bf sFID$\downarrow$ & \bf FID$\downarrow$ \\
\midrule
($\mathcal{N}$=3, $\mathcal{O}$=0) & 8.56 & 2.77$\times$ & 6.36 & 3.55 \\
($\mathcal{N}$=4, $\mathcal{O}$=0) & 6.66 & 3.56$\times$ & 8.43 & 4.75 \\
($\mathcal{N}$=5, $\mathcal{O}$=0) & 5.24 & 4.53$\times$ & 11.29 & 6.58 \\
($\mathcal{N}$=6, $\mathcal{O}$=0) & 4.76 & 4.98$\times$ & 14.84 & 9.24 \\
($\mathcal{N}$=7, $\mathcal{O}$=0) & 3.82 & 6.22$\times$ & 18.57 & 12.67 \\
\midrule
($\mathcal{N}$=3, $\mathcal{O}$=1) & 8.56 & 2.77$\times$ & 4.82 & 2.49 \\
($\mathcal{N}$=4, $\mathcal{O}$=1) & 6.66 & 3.56$\times$ & 5.54 & 2.66 \\
($\mathcal{N}$=5, $\mathcal{O}$=1) & 5.24 & 4.53$\times$ & 6.48 & 3.05 \\
($\mathcal{N}$=6, $\mathcal{O}$=1) & 4.76 & 4.98$\times$ & 7.62 & 3.59 \\
($\mathcal{N}$=7, $\mathcal{O}$=1) & 3.82 & 6.22$\times$ & 8.76 & 4.29 \\
\midrule
($\mathcal{N}$=3, $\mathcal{O}$=2) & 8.56 & 2.77$\times$ & 4.69 & 2.44 \\
($\mathcal{N}$=4, $\mathcal{O}$=2) & 6.66 & 3.56$\times$ & 5.21 & 2.51 \\
($\mathcal{N}$=5, $\mathcal{O}$=2) & 5.24 & 4.53$\times$ & 5.87 & 2.79 \\
($\mathcal{N}$=6, $\mathcal{O}$=2) & 4.76 & 4.98$\times$ & 6.65 & 3.18 \\
($\mathcal{N}$=7, $\mathcal{O}$=2) & 3.82 & 6.22$\times$ & 7.26 & 3.66 \\
\midrule
($\mathcal{N}$=3, $\mathcal{O}$=3) & 8.56 & 2.77$\times$ & 4.69 & 2.34 \\
($\mathcal{N}$=4, $\mathcal{O}$=3) & 6.66 & 3.56$\times$ & 5.21 & 2.53 \\
($\mathcal{N}$=5, $\mathcal{O}$=3) & 5.24 & 4.53$\times$ & 5.82 & 2.78 \\
($\mathcal{N}$=6, $\mathcal{O}$=3) & 4.76 & 4.98$\times$ & 6.50 & 3.15 \\
($\mathcal{N}$=7, $\mathcal{O}$=3) & 3.82 & 6.22$\times$ & 7.08 & 3.63 \\
\midrule
($\mathcal{N}$=3, $\mathcal{O}$=4) & 8.56 & 2.77$\times$ & 4.69 & 2.35 \\
($\mathcal{N}$=4, $\mathcal{O}$=4) & 6.66 & 3.56$\times$ & 5.19 & 2.52 \\
($\mathcal{N}$=5, $\mathcal{O}$=4) & 5.24 & 4.53$\times$ & 5.82 & 2.78 \\
($\mathcal{N}$=6, $\mathcal{O}$=4) & 4.76 & 4.98$\times$ & 6.50 & 3.15 \\
($\mathcal{N}$=7, $\mathcal{O}$=4) & 3.82 & 6.22$\times$ & 7.07 & 3.63 \\
\bottomrule
\end{tabularx}

\end{table}

%% file: tab/HiDream-Metrics.tex
\begin{table}[h]
\centering
\caption{Performance comparison on HiDream.}
\vspace{-3mm}
\setlength\tabcolsep{2pt}
\renewcommand{\arraystretch}{1.2} 
\tiny
\begin{tabular}{l|cc|c|ccc}
\toprule
{\bf Method} & \multicolumn{2}{c|}{\bf Acceleration} & {\bf Image} & \multirow{2}{*}{\bf PSNR$\uparrow$} & \multirow{2}{*}{\bf SSIM$\uparrow$} & \multirow{2}{*}{\bf LPIPS$\downarrow$} \\
\cline{2-3}
 {\textbf{HiDream}}& {\bf TFLOPs$\downarrow$} & {\bf Speed$\uparrow$} & {\bf Reward$\uparrow$} & & & \\
\midrule
\textbf{HiDream-Full} & 7780.0 & 1.0$\times$ & 1.1285 & - & - & - \\
\textbf{TeaCache}($l_1 = 1$) & 2047.4 & 3.8$\times$ & 0.9849 & 28.139 & 0.6036 & 0.565 \\
\rowcolor{gray!20}
\textbf{\textit{TaylorSeer}}($N = 4, O = 2$) & \textbf{1945.0} & \textbf{4.0}$\times$ & \textbf{1.0833} & \textbf{28.248} & \textbf{0.6084} & \textbf{0.532 }\\
\bottomrule
\end{tabular}
\label{table:hidream-comparison-eng}
\end{table}